\documentclass[11pt]{article}

\usepackage[final]{acl}

\usepackage{times}
\usepackage{graphicx}
\usepackage{latexsym}
\usepackage{amsmath}
\usepackage{amssymb}
\usepackage[T1]{fontenc}
\usepackage[utf8]{inputenc}
\usepackage{microtype}
\usepackage{inconsolata}
\usepackage{algorithm}
\usepackage{algpseudocode}
\usepackage{xcolor}

\usepackage{booktabs}
\usepackage{tabularx}
\usepackage{caption}
\usepackage{float}      
\usepackage{placeins}     
\usepackage{dblfloatfix}  
\newcommand{\VTI}{\textsc{VTI}}
\definecolor{vlblue}{RGB}{80,160,220}

\algrenewcommand\algorithmiccomment[1]{\hfill{\color{vlblue}\footnotesize$\triangleright$~#1}}
\algrenewcommand\algorithmicrequire{\textbf{Input:}}
\algrenewcommand\algorithmicensure{\textbf{Output:}}

\algnewcommand\algorithmicparfor{\textbf{parallel for}}
\algdef{S}[FOR]{ParFor}[1]{\algorithmicparfor\ #1}
\algdef{E}[FOR]{EndParFor}{\algorithmicend\ \algorithmicparfor}

\newcommand{\FlashOS}{\operatorname{FlashOS}}
\newcommand{\ReduceSoftmax}{\operatorname{ReduceSoftmax}}

\title{HyQuant: Hybrid-Precision Quantization for LLM Attention}

\author{
  \textbf{Jiatong Ding}\textsuperscript{1*},
  \textbf{Bingxin Xing}\textsuperscript{2*},
  \textbf{Yu Zhang}\textsuperscript{4},
  \textbf{Dian Ding}\textsuperscript{1\dag},
  \textbf{Xiaodong Yi}\textsuperscript{3},
  \textbf{Xianbin Ouyang}\textsuperscript{3}, \\
  \textbf{Feihu Zhou}\textsuperscript{3},
  \textbf{Kun Zhang}\textsuperscript{3},
  \textbf{Zhenyu Guo}\textsuperscript{3},
  \textbf{Hao Pan}\textsuperscript{1},
  \textbf{Guangtao Xue}\textsuperscript{1},
  \textbf{Yiming Zhang}\textsuperscript{1} \\
  \normalfont\normalsize
  \textsuperscript{1}Shanghai Jiao Tong University,
  \textsuperscript{2}Xi'an Jiaotong University,
  \textsuperscript{3}Tencent Penglai Lab,
  \textsuperscript{4}Xiamen University
}


\begin{document}
\maketitle
\begingroup
  \renewcommand\thefootnote{}%
  \footnotetext{\textsuperscript{*}\,Equal contribution.\quad
                \textsuperscript{\dag}\,Corresponding author.}%
\endgroup
\begin{abstract}
Quantization has been widely adopted in LLM training and inference to reduce cost and improve efficiency.
However, low-bit quantization of the \emph{attention} module often introduces large errors at very low bit-widths, causing performance degradation.
Existing methods mainly rely on smoothing techniques to handle outliers, while we propose a hybrid quantization design to better balance accuracy and efficiency. Specifically, we propose \textbf{HyQuant}, an efficient hybrid quantization framework for LLM attention.
HyQuant quantizes most attention states into low-bit formats while retaining a small set of vertical-line tokens and local-window states in high precision.
These accuracy-critical regions are selected using lightweight vertical-line-aware attention-pattern signals, reducing quantization error with limited overhead. In the Prefill stage, HyQuant uses a hybrid-precision quantized attention operator that preserves vertical-line tokens and a local sliding window in full precision while quantizing the remaining context. In the Decode stage, HyQuant applies the same principle to KV-cache compression and fuses KV dequantization with attention computation to improve memory and hardware efficiency. Across diverse tasks, models, and datasets, HyQuant maintains nearly lossless
accuracy with an extremely simple design, demonstrating the efficiency and
practical feasibility of hybrid quantization for LLM attention. Code is available at: \url{https://github.com/jerrysfls/HyQuant}.
\end{abstract}

\section{Introduction}
With OpenAI's o1 series~\cite{openai2024o1}, Gemini~\cite{gemini}, and DeepSeek~\cite{deepseek2025r1} bringing Chain-of-Thought (CoT) reasoning into mainstream, Large Language Models (LLMs) have shifted from brief answers to long multi-step reasoning traces, often reaching tens of thousands of tokens.

In long-context inference, bottlenecks differ across stages.
During \emph{Prefill}, full-prefix attention is dominated by large matrix multiplications and is mainly compute-bound.
During \emph{Decode}, each new token repeatedly reads/writes past KV states, so memory capacity and bandwidth become the constraints.

Quantization is widely deployed to reduce compute and memory costs, but
aggressively pushing precision to very low bit-widths can degrade
end-to-end model quality
~\cite{zheng2025empiricalstudyqwen3quantization}.
In our setting, uniform token-wise precision further fails to account for
the non-uniform sensitivity induced by imbalanced attention.
This stems from \emph{non-uniform sensitivity} under imbalanced attention, a few high-contribution positions dominate the error, so uniform compression over-compresses critical tokens while wasting budget elsewhere.
Existing approaches, whether sparsification-based (eviction, block-sparse attention) or quantization-based (KV-cache quantization, approximated attention) largely treat tokens as homogeneous and \textbf{do not answer a reasoning-centric mixed-precision question}: how to design a mixed precision that accounts for heterogeneous token sensitivity to balance accuracy and efficiency in long-context CoT inference. 

\begin{figure}[t]
\centering
\setlength{\tabcolsep}{1pt}
\renewcommand{\arraystretch}{0.4}
\begin{tabular}{@{}c@{\hspace{2pt}}ccc@{}}
\rotatebox[origin=c]{90}{\tiny\textbf{Qwen3-8B}} &
\includegraphics[width=0.3\columnwidth]{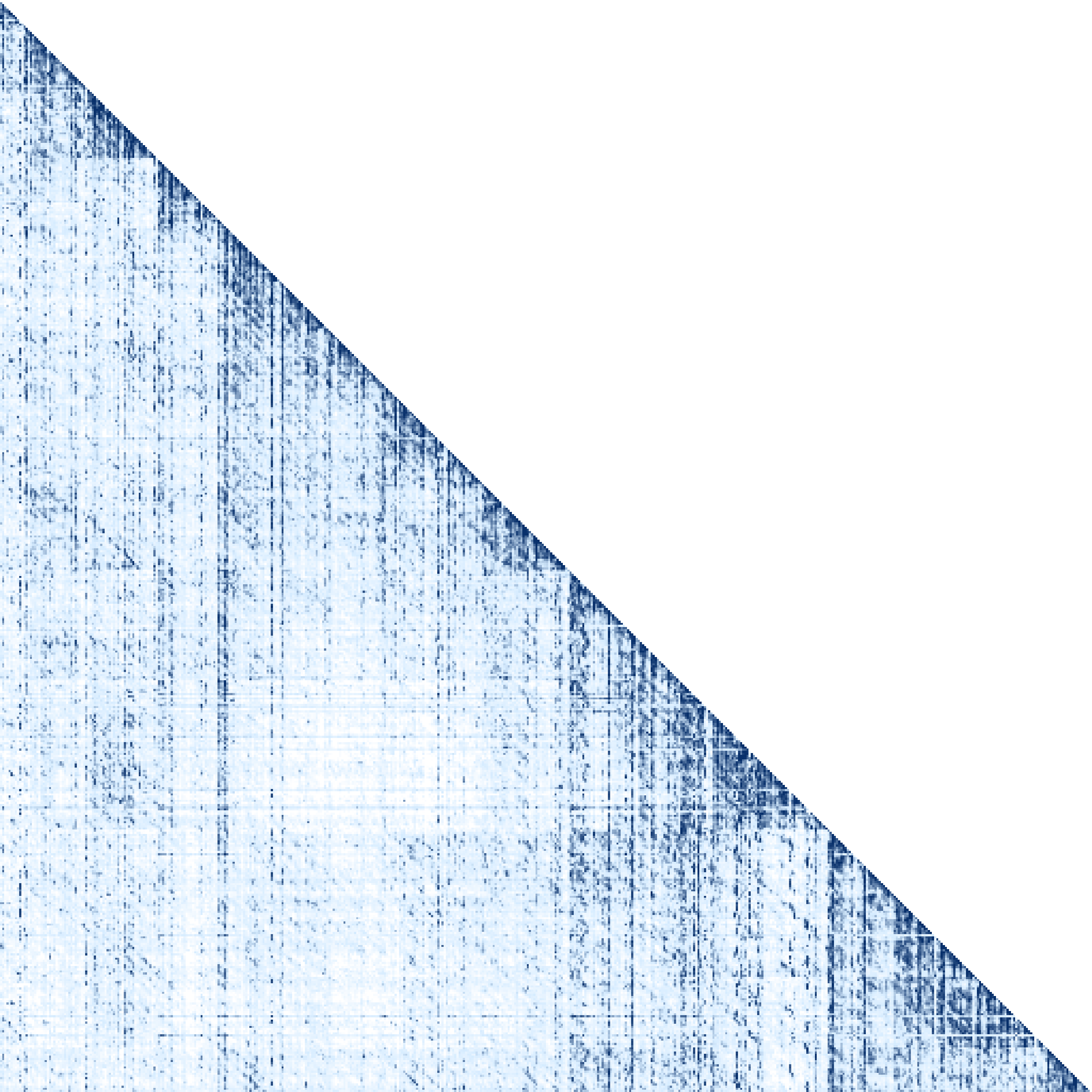} &
\includegraphics[width=0.3\columnwidth]{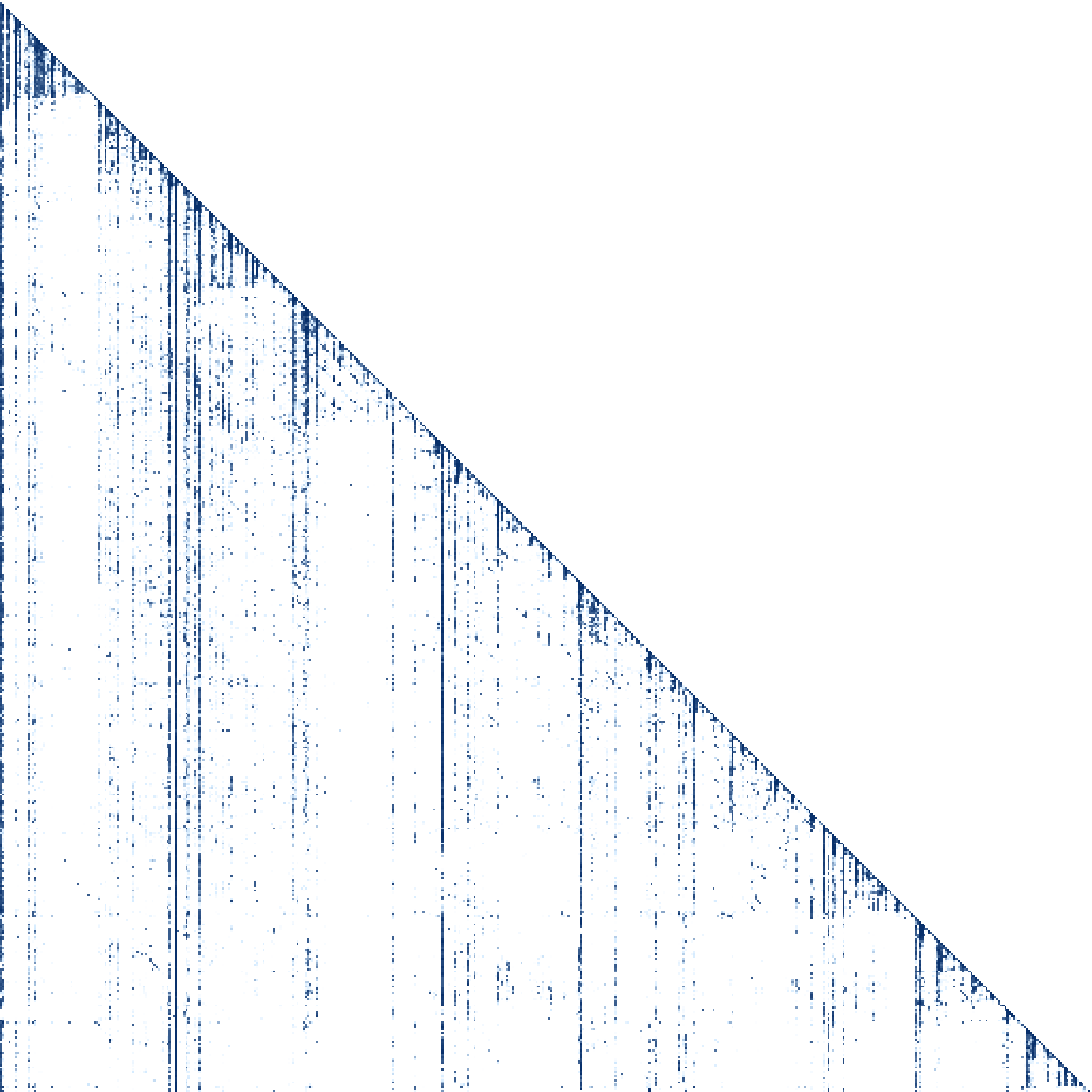} &
\includegraphics[width=0.3\columnwidth]{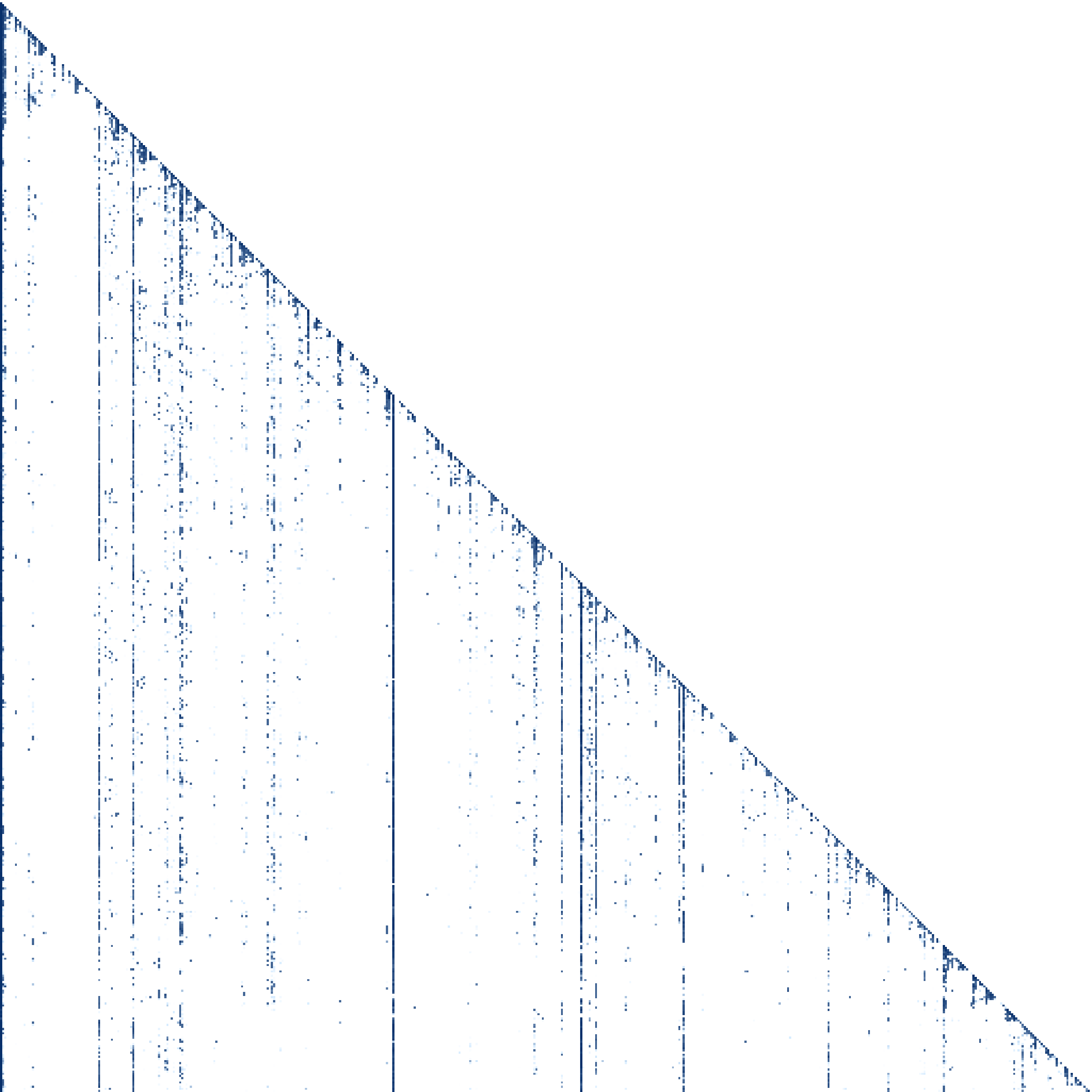} \\
\rotatebox[origin=c]{90}{\tiny\textbf{Gemma4-31B}} &
\includegraphics[width=0.3\columnwidth]{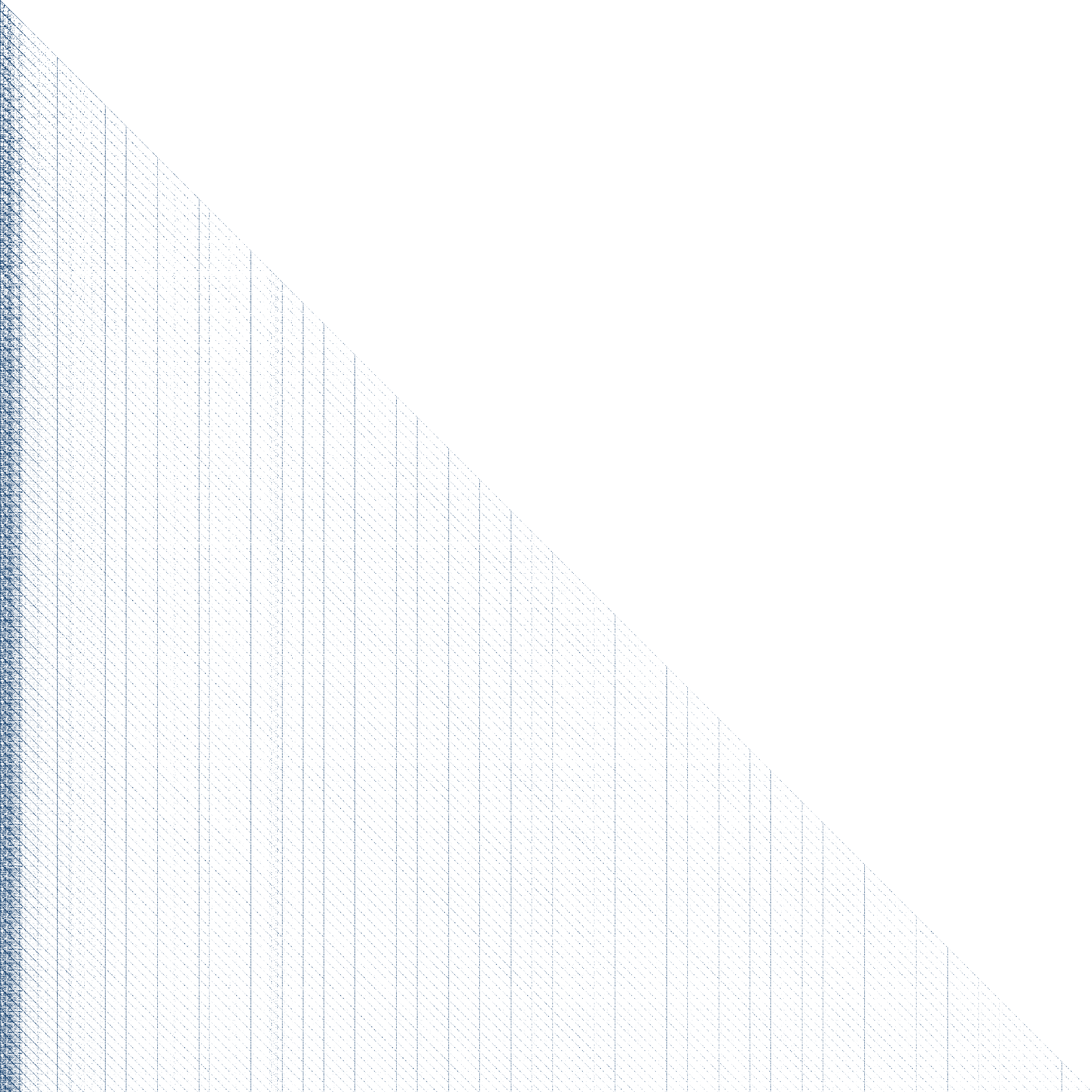} &
\includegraphics[width=0.3\columnwidth]{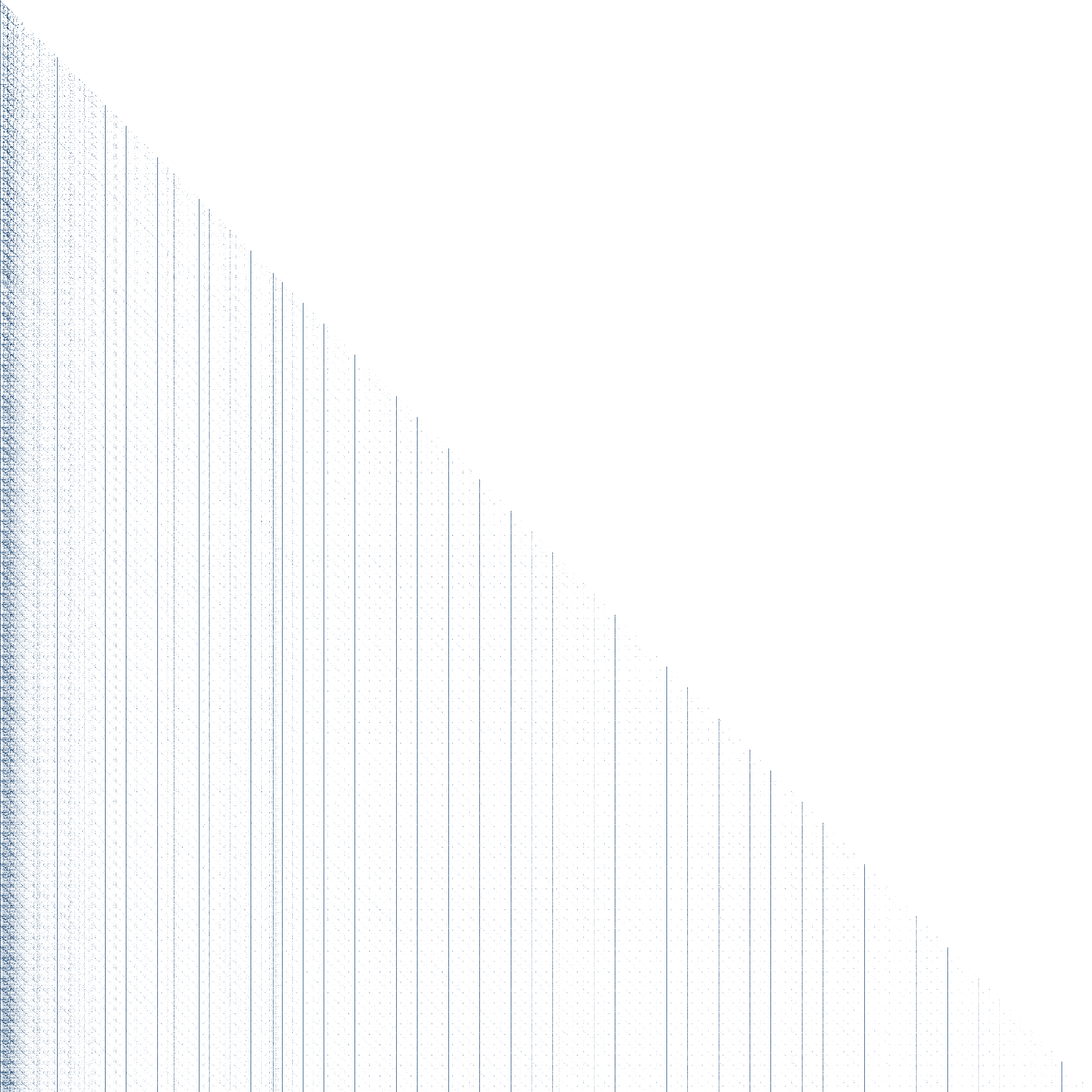} &
\includegraphics[width=0.3\columnwidth]{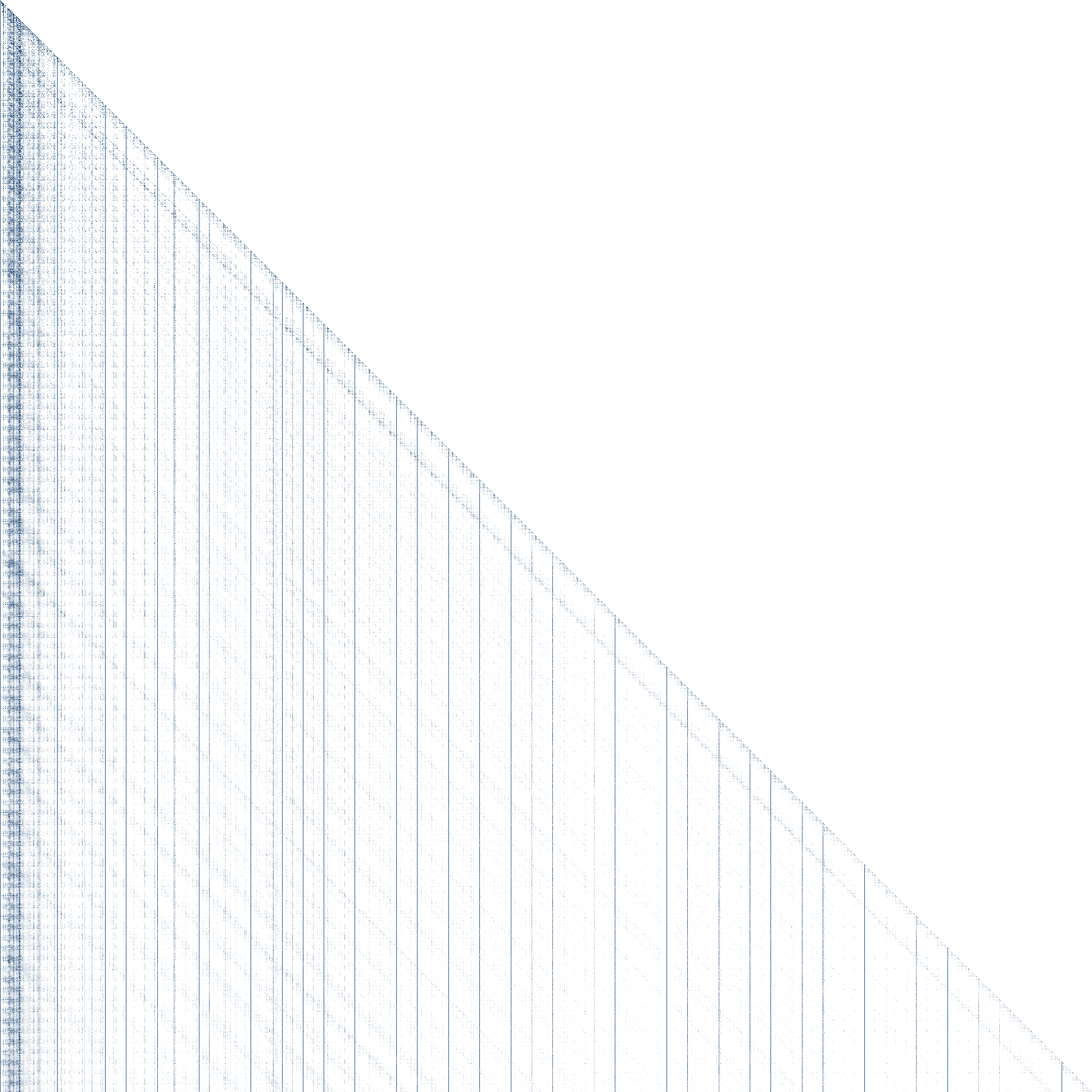} \\
\rotatebox[origin=c]{90}{\tiny\textbf{Qwen3.5-4B}} &
\includegraphics[width=0.3\columnwidth]{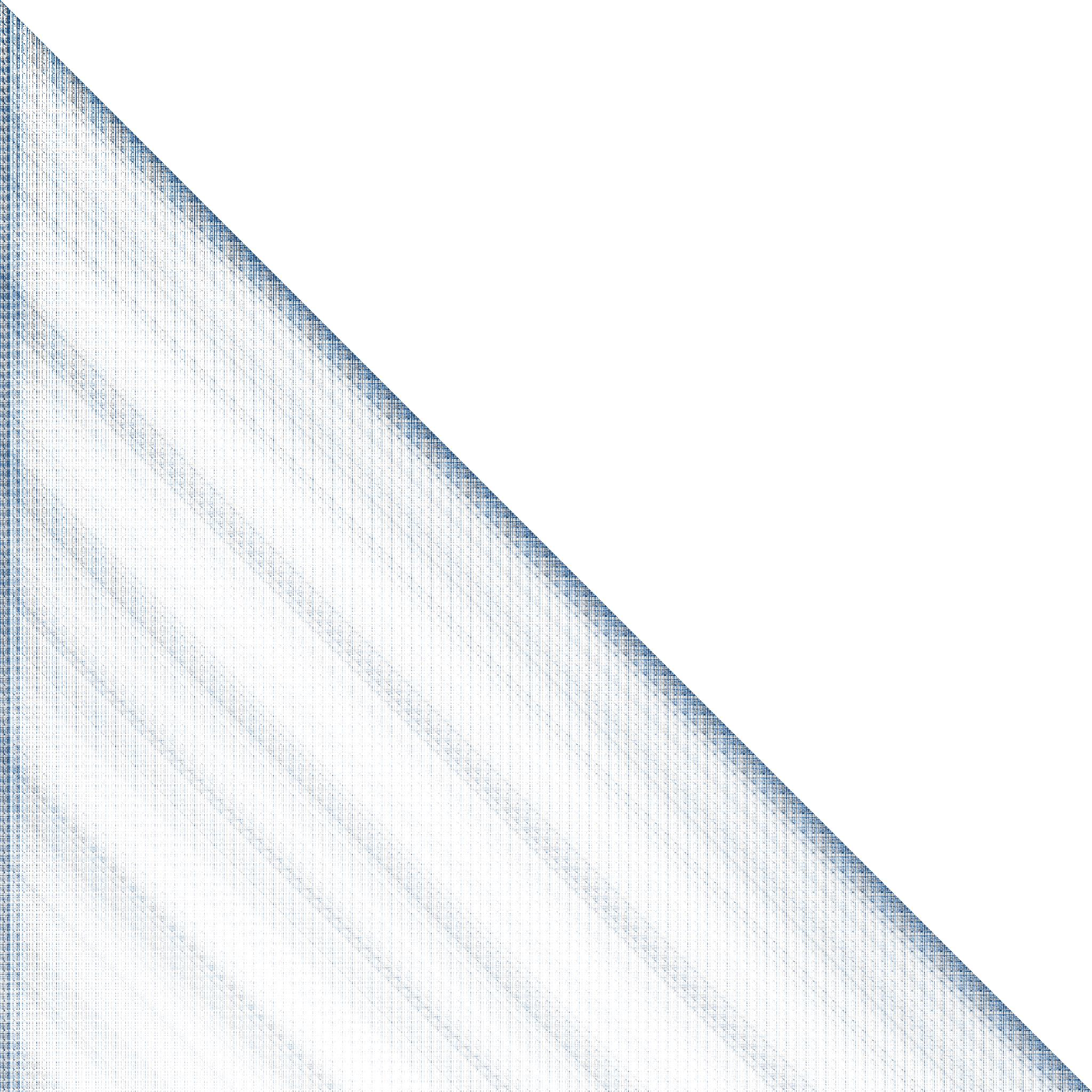} &
\includegraphics[width=0.3\columnwidth]{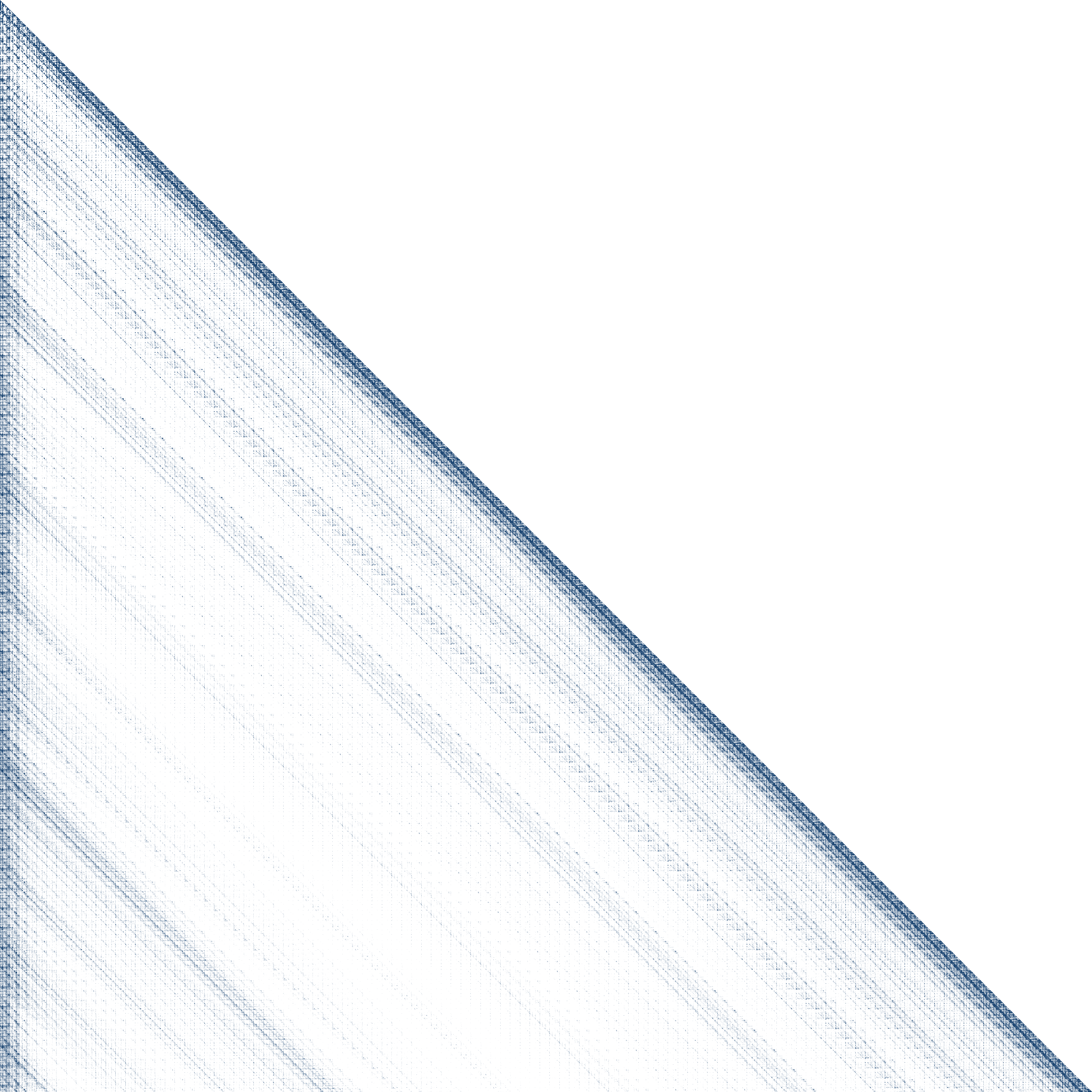} &
\includegraphics[width=0.3\columnwidth]{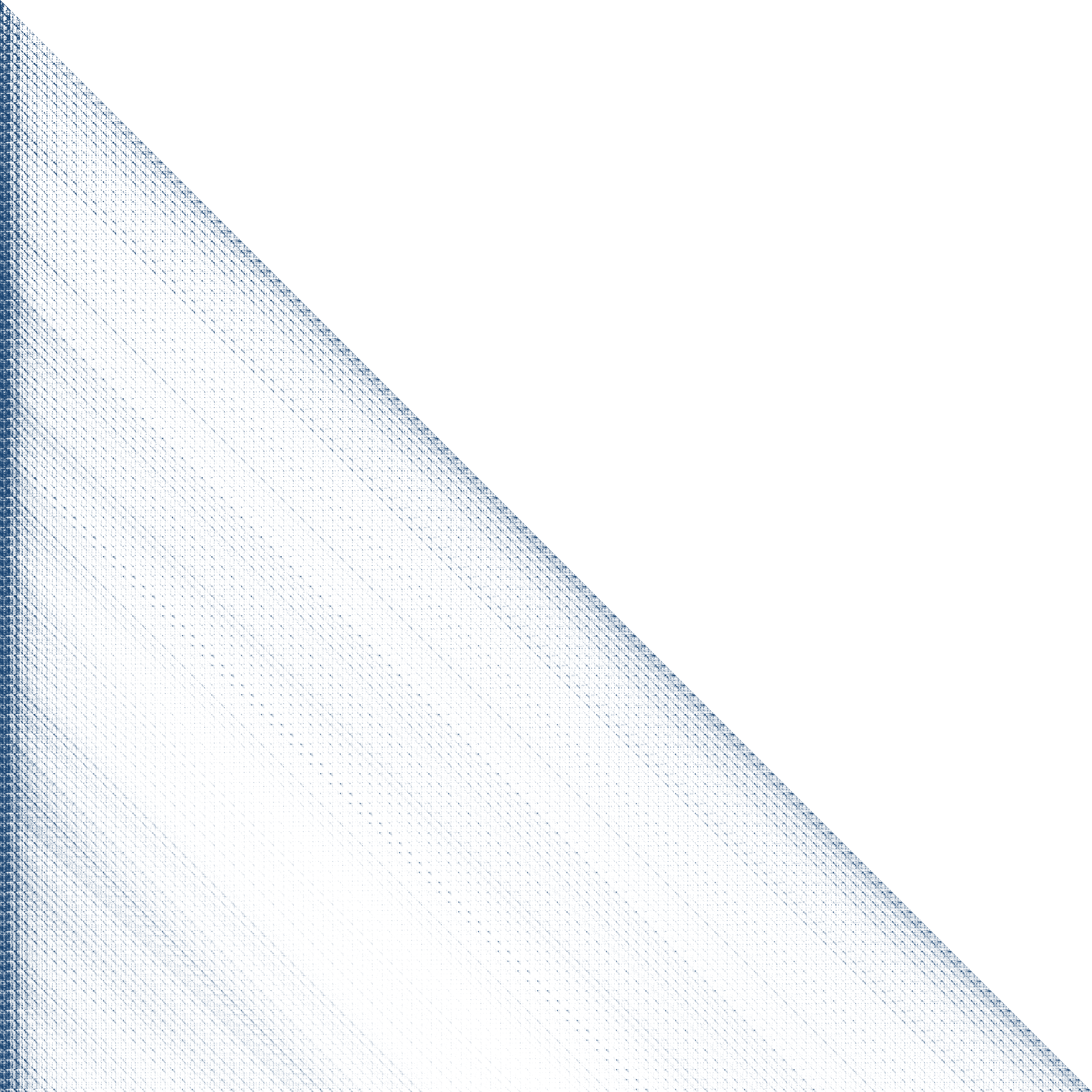} \\
\rotatebox[origin=c]{90}{\tiny\textbf{Llama3-8B}} &
\includegraphics[width=0.3\columnwidth]{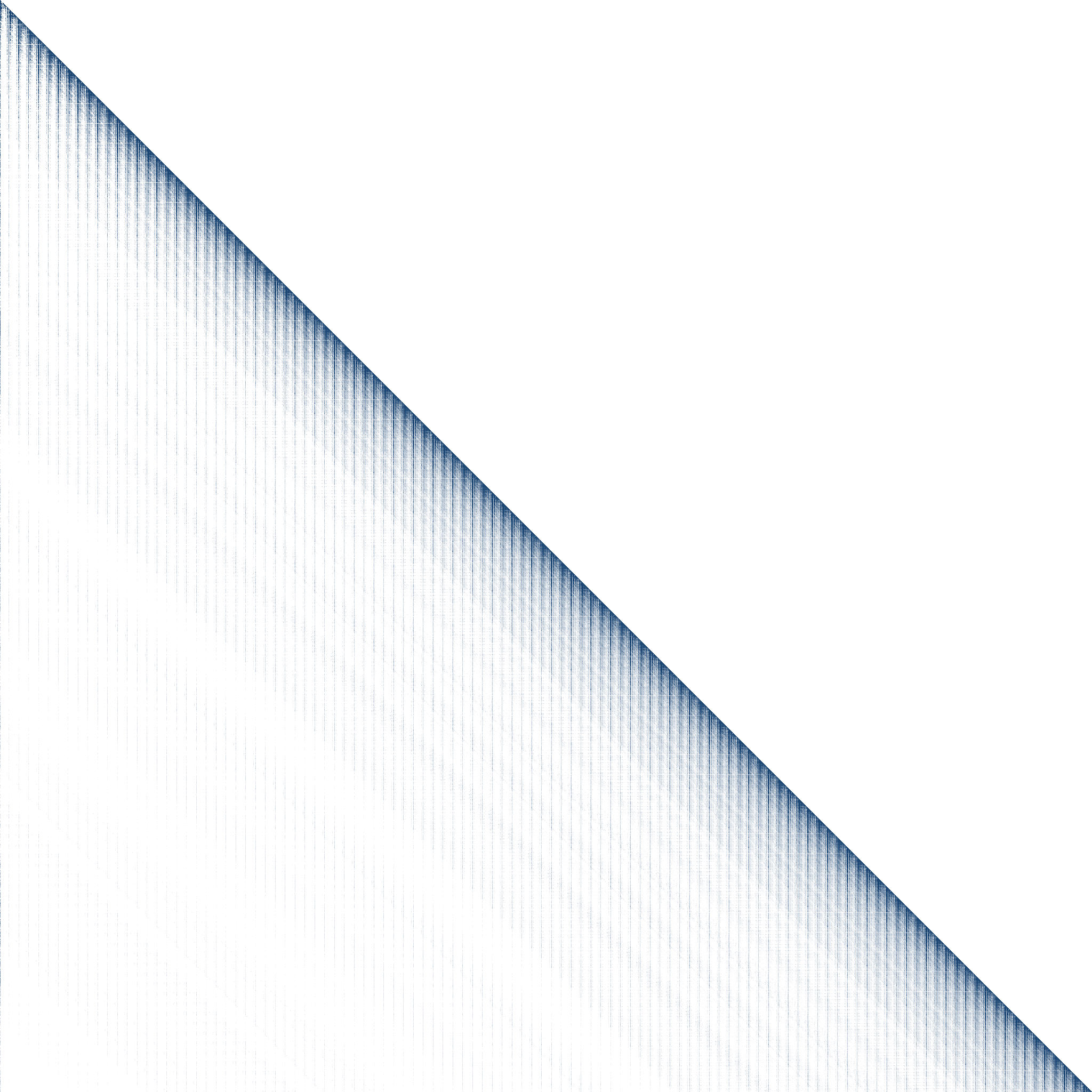} &
\includegraphics[width=0.3\columnwidth]{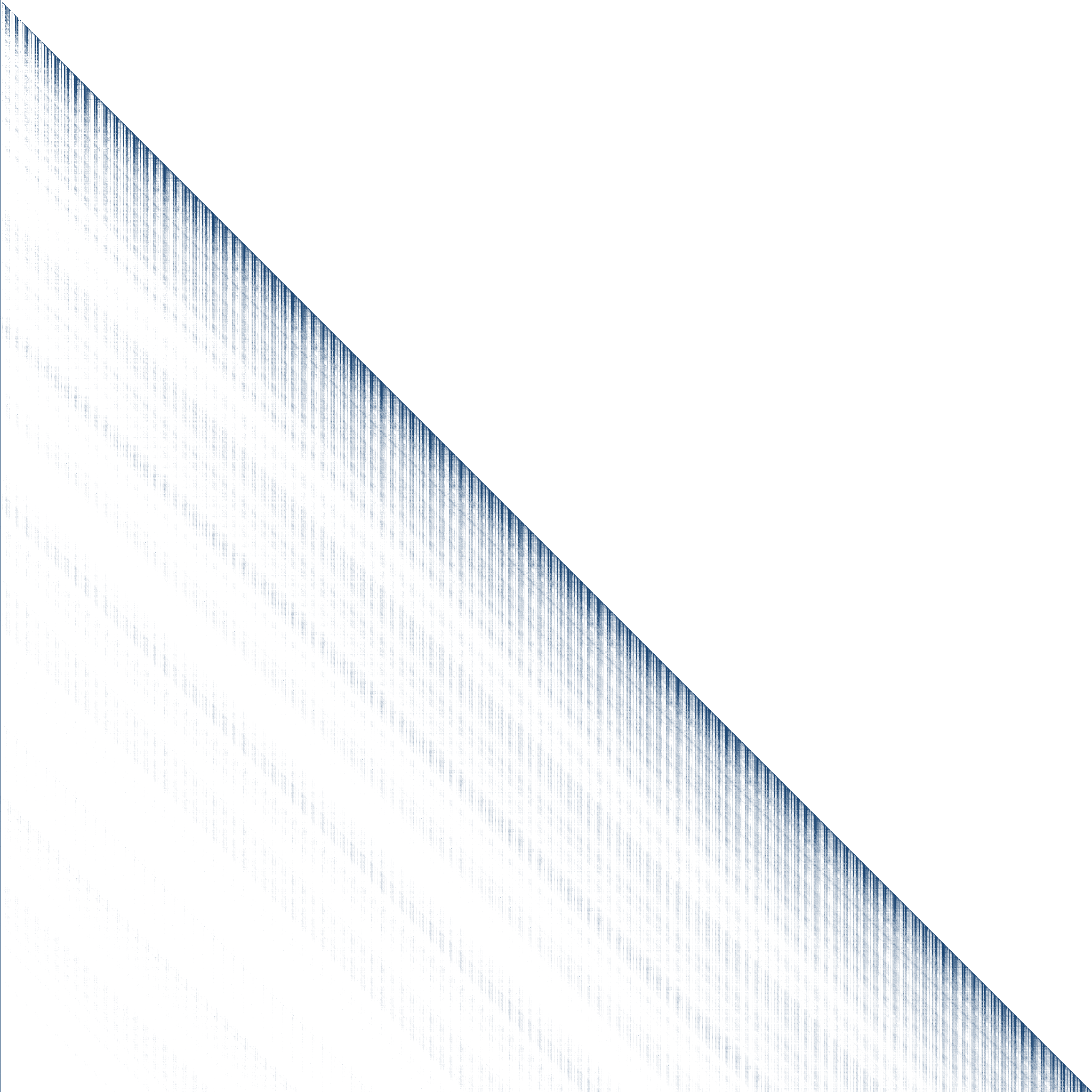} &
\includegraphics[width=0.3\columnwidth]{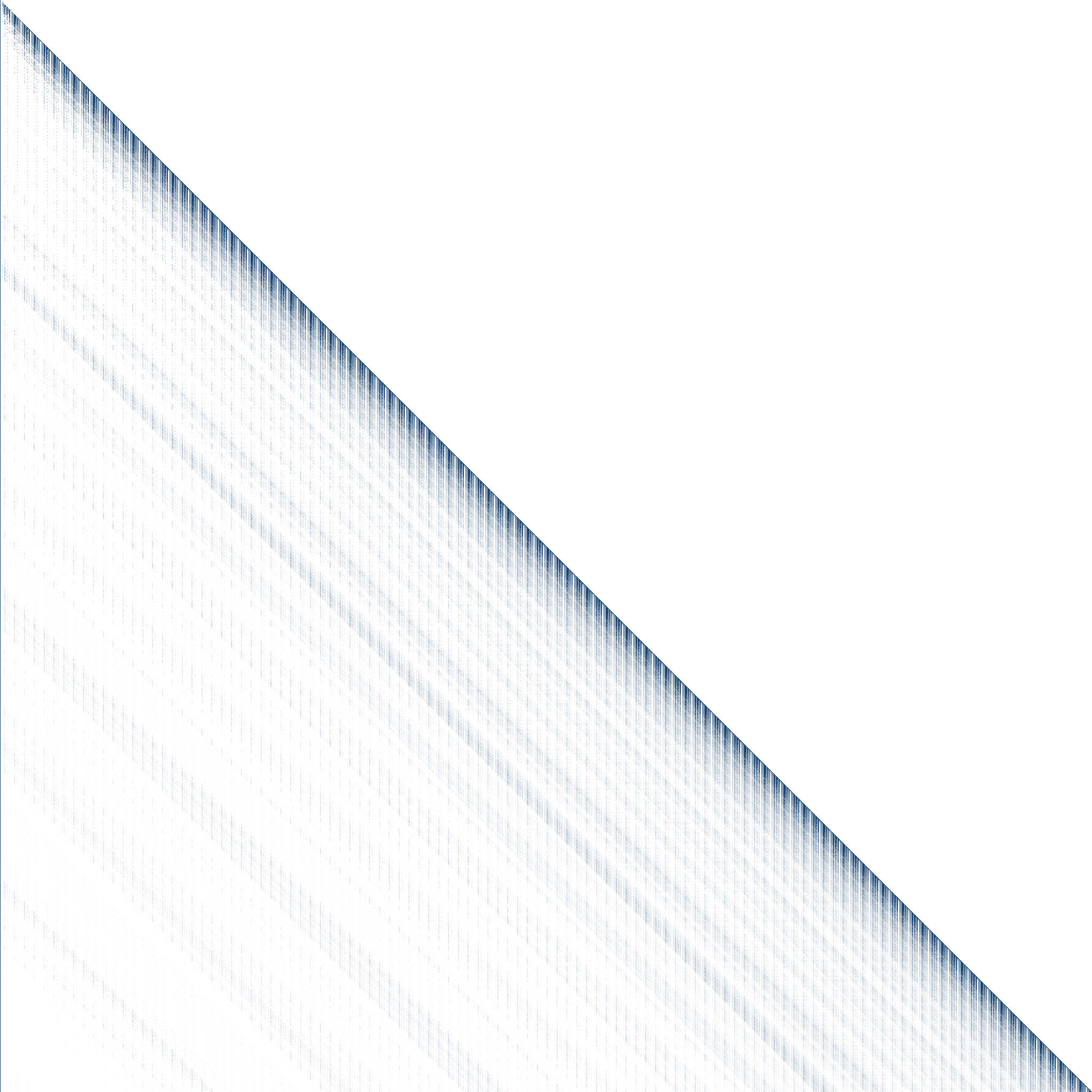} \\
\end{tabular}
\caption{\textbf{Vertical-line-dominated attention across model families.}
Rows from top to bottom: Qwen3-8B, Gemma4-31B, Qwen3.5-4B, and Llama3-8B.
Across layers, heads, and model families, persistent vertical bright stripes indicate that a small number of positions are repeatedly attended over long spans, revealing a highly imbalanced attention distribution and motivating compression strategies that prioritize a small set of critical positions.}
\label{fig:vertical_heatmaps_singlecol}
\end{figure}

As shown in Fig.~\ref{fig:vertical_heatmaps_singlecol}, we repeatedly observe \emph{vertical-line} structures across attention heatmaps, consistent with prior observations~\cite{jiang2024minference10acceleratingprefilling}.
Unlike drifting diagonal and oblique bands, these vertical lines carry higher and more stable attention mass, but cover only a tiny fraction of tokens (typically $<5\%$).
This motivates preserving such positions in full precision during low-bit quantization.

Motivated by these observations, we propose \textbf{HyQuant}, a \textbf{mixed-precision quantization} method that keeps full precision for a tiny set of positions persistent \emph{vertical-line} positions and a recent \emph{sliding window} while quantizing the rest to low bit-width.
HyQuant identifies vertical lines with a lightweight procedure and uses a fixed window size with small overhead.
In \textbf{Prefill} (compute-intensive), we design a quantized attention operator that runs most computation in low precision while keeping vertical lines and the local window in full precision, fusing both paths into a single operator.
In \textbf{Decode} (memory-intensive), we quantize the KV cache to reduce capacity and bandwidth pressure while keeping the same critical positions unquantized, and fuse KV dequantization with attention computation for efficiency.

We evaluate \textbf{HyQuant} on Qwen3-8B, Qwen3-32B, LLaMA3.1-8B, and GLM-4-9B with LongBench, GSM8K, and MATH500, achieving \textbf{1.32$\times$ to 3.58$\times$} decode-kernel speedup and \textbf{1.04$\times$ to 1.17$\times$} end-to-end decode speedup, while maintaining near-full-precision accuracy and improving over strict low-bit baselines in most settings.

\section{Motivation}
\label{sec:motivation}

\subsection{A Small Number of Tokens Remain Persistently Important}
\label{sec:motivation:vertical}

HyQuant is motivated by a recurring observation in real inference: LLM attention
is highly non-uniform, with a large fraction of attention mass concentrated on a
small set of salient tokens. Similar structures have been reported in prior
long-context studies, including the vertical-line and vertical-slash patterns in
MInference~\cite{jiang2024minference10acceleratingprefilling} and the attention
sink phenomenon in StreamingLLM~\cite{xiao2024streamingllm}. Although these
patterns differ in form, they reveal the same underlying property: only a small
subset of key/value positions are repeatedly attended to across long generation
spans. These positions act as persistent attention anchors. Depending on the
pattern, they may correspond either to sink-like special positions or
to semantically relevant tokens that are repeatedly referenced by
later queries.

Fig.~\ref{fig:vertical_heatmaps_singlecol} shows that this vertical-stripe
pattern is pervasive within Qwen3-8B. Beyond the local diagonal structure, many
layers and heads exhibit bright vertical columns that are attended by many query
tokens, indicating that the \emph{effective attention support} is concentrated
rather than uniformly distributed over the full sequence. The same figure further shows that this phenomenon is not limited to Qwen3-8B: recent model families, including Gemma4-31B, Qwen3.5-4B, and Llama3-8B, exhibit similar sink-like vertical patterns.

This observation is consistent with prior analyses of attention sinks
\cite{xiao2024streamingllm,gu2025attentionsinkemergeslanguage}. Recent gated-attention mechanisms have
been proposed to mitigate such sink effects~\cite{qiu2025gatedattentionlargelanguage}, but our
heatmaps suggest that sink-like vertical concentration is still not fully removed
in modern models. Therefore, token-agnostic KV eviction or uniform low-bit
quantization can be mismatched with real long-context inference dynamics,
motivating hybrid compression methods that explicitly preserve these
persistently attended tokens.

\paragraph{Evidence: a small set of high-score positions covers most attention mass.}
We further quantify this concentrated-attention phenomenon on \textbf{Llama-3.1-8B} and \textbf{Qwen3-8B}.
For each model, we measure the fraction of attention mass, computed from softmax$(QK^\top)$, covered by:
(i) the global top-$1\%$ highest-scoring key positions (\textsc{Top-1\%}),
(ii) the global top-$5\%$ highest-scoring key positions (\textsc{Top-5\%}), and
(iii) the union of the global top-$5\%$ key positions and a local window $W{=}128$ (\textsc{Top-5\%+Win}).
As shown in Table~\ref{tab:mass_topk_llama_gemma}, a small fraction of high-score key positions already covers a substantial portion of the total attention mass, and combining them with a short local window further increases coverage to over $80\%$.
This confirms that the effective attention support is dominated by a small set of persistently important tokens together with recent local context, motivating a \emph{non-uniform} precision allocation that aligns the precision budget with this concentrated structure.

\begin{table}[t]
  \centering
  \small
  \caption{\textbf{Attention-mass coverage across model families.}
  We report the fraction of attention mass covered by the global top-$1\%$ key positions
  (\textsc{Top-1\%}), the global top-$5\%$ key positions (\textsc{Top-5\%}),
  and the union of the global top-$5\%$ key positions and a local window $W{=}128$
  (\textsc{Top-5\%+Win}).}
  \label{tab:mass_topk_llama_gemma}

  \setlength{\tabcolsep}{4pt}
  \renewcommand{\arraystretch}{1.15}

  \begin{tabularx}{\columnwidth}{lccc}
    \toprule
    Model & \textsc{Top-1\%} & \textsc{Top-5\%} & \textsc{Top-5\%+Win} \\
    \midrule
    \textbf{Llama-3.1-8B} & 58.83\% & 64.09\% & 85.63\% \\
    \textbf{Qwen3-8B}     & 47.30\% & 54.10\% & 82.53\% \\
    \bottomrule
  \end{tabularx}
\end{table}

\subsection{Quantization Errors Are Amplified at High-Score Positions}
\label{sec:motivation:quant_error}

Low-bit quantization errors are not equally harmful across tokens: they are most destructive at \emph{high-score} positions, where small K/V perturbations get amplified through repeated KV access by many queries across prefill and decode.
As a result, uniform quantization with a fixed compression rate often fails to retain a small number of critical tokens in sufficient precision, leading to considerable quality degradation that is consistent with the attention imbalance observed in Section~\ref{sec:motivation:vertical}.

\begin{figure}[t]
  \centering
  \includegraphics[width=0.90\linewidth]{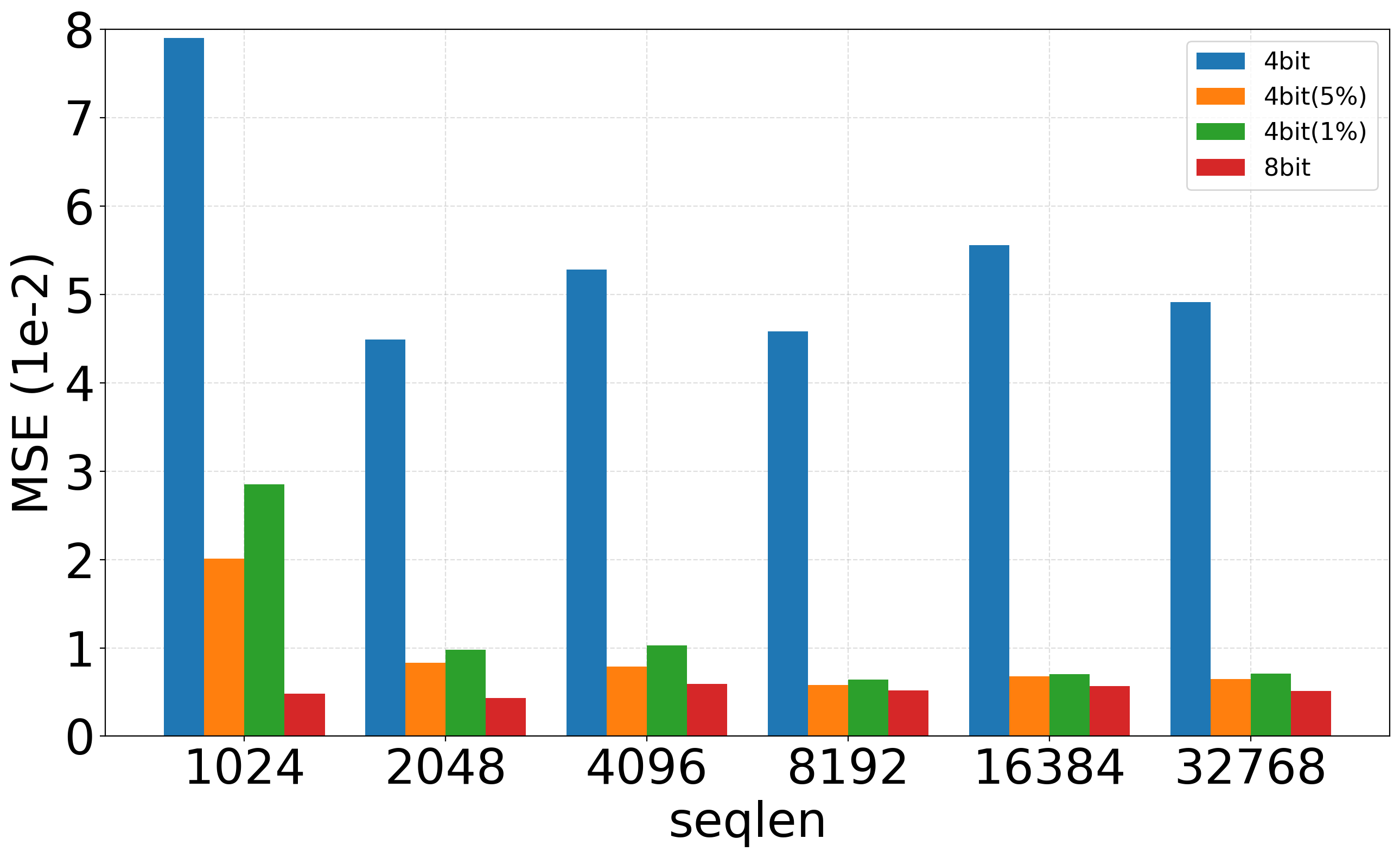}
  \caption{\textbf{Amplified quantization error at high-score positions (Qwen3-8B, Layer~28).}
  Uniform 4-bit quantization incurs substantially larger MSE than uniform 8-bit quantization.
  When the remaining positions are quantized to 4-bit, retaining only the top-$1\%$/top-$5\%$ high-score positions in full precision significantly reduces the error
  and consistently approaches the 8-bit error level across different sequence lengths.}
  \label{fig:2}
\end{figure}

\paragraph{Evidence: retaining a tiny set of high-score positions in full precision substantially reduces error.}
We conduct error analysis on Qwen3-8B and use full-precision FlashAttention as the reference, measuring the MSE of the intermediate attention output (input to \texttt{o\_proj}) under different quantization strategies.
As shown in Fig.~\ref{fig:2}, uniform 4-bit quantization yields noticeably larger errors than uniform 8-bit quantization.
More importantly, when the remaining positions are quantized to 4-bit, retaining only the top-$1\%$/top-$5\%$ high-score positions in full precision dramatically suppresses the error—consistently approaching the 8-bit error level across sequence lengths from 1K to 32K.
This directly motivates the hybrid-precision design of HyQuant: with a small full-precision budget, HyQuant prioritizes critical high-attention K/V positions while quantizing the remaining majority to low bit-width.

\section{Related Work}
\label{sec:related_work}

\subsection{Sparse and Selective Long-Context Inference}
A major line of work accelerates long-context inference by skipping
low-contribution attention blocks or selectively retaining cache content.
For \emph{Prefill}, block-sparse attention reduces FLOPs by computing
only selected blocks, often relying on online scoring, retrieval, or
reordering. These methods are effective at very long contexts, but their
preprocessing overhead can offset the saved FLOPs at short or moderate
sequence lengths. For \emph{Decode}, KV-cache eviction and selective
retention methods such as H2O~\cite{zhang2023h2oheavyhitteroracleefficient},
SnapKV~\cite{li2024snapkvllmknowslooking},
PyramidKV~\cite{cai2025pyramidkvdynamickvcache}, and
HeadKV~\cite{fu2025headsmatterheadlevelkv} reduce memory footprint.
However, eviction can be brittle for long CoT reasoning, where token
importance is non-stationary and previously removed content may later
become critical. In contrast, the vertical-line phenomenon we exploit is measured at the query-distribution level: certain key positions receive high attention from a large fraction of query positions rather than from only a single query. Similar persistent concentration patterns have been reported in analyses of dynamic sparse attention and attention sinks \citep{jiang2024minference10acceleratingprefilling,xiao2024streamingllm}.

\subsection{Low-Bit Attention and KV-Cache Quantization}
Quantization reduces bandwidth and storage via low-bit representations.
In \emph{Prefill}, quantized or approximated attention methods such as
SageAttention~\cite{zhang2025sageattentionaccurate8bitattention} improve
attention efficiency, but aggressive bit-widths can introduce noticeable
quality loss. In \emph{Decode}, KV-cache quantization is more mature:
KIVI~\cite{liu2024kivi}, KVTuner~\cite{li2025kvtuner}, and
KVQuant~\cite{hooper2024kvquant} reduce quantization error through
asymmetric quantization, sensitivity-aware allocation, or outlier handling.
These methods mainly focus on reducing quantization error under compact
representations, but they do not explicitly exploit persistent
vertical-line structures for token-level full-precision retention in
long-context reasoning.

\subsection{Difference from previous work.} A closely related work, exemplified by MInference \cite{jiang2024minference10acceleratingprefilling}, also identifies vertical-line structures in attention maps. However, unlike MInference, which employs these patterns as a sparsity mask, we use them to assign different precisions to tokens of various importance. This method retains information from less significant tokens in a relatively low bit form, which is totally overlooked in MInference. To visualize the effect in long-context settings, we conduct an ablation study on Qwen3-8B and Longbench dataset, whose results can be seen in Section \ref{sec:conceptablation}. In addition, the idea of sensitivity-aware mixed-precision appeared in KVTuner \cite{li2025kvtuner}, yet the author applied it to different layers rather than tokens in our work. 

Another highlight in our work is that we propose an integrated Prefill-Decode method, providing optimization to both ends, while previous methods usually deal with only one stage According to identified vertical line tokens, we implement operand quantization in prefill stage, and both operand and cache quantization in decode stage. This results in both accelerated speed, decreased memory usage, and increased accuracy.
\section{Method}
\label{sec:method}

We present HyQuant, a vertical-line-aware hybrid-precision quantization framework for long-context inference.
As illustrated in Fig.~\ref{fig:vertical_heatmaps_singlecol}, a tiny fraction of tokens (typically $<5\%$) forms persistent \emph{vertical lines} that carry disproportionately large attention mass and dominate quantization error under aggressive low-bit settings.
HyQuant retains these error-sensitive vertical-line positions and a local window in full precision, while quantizing the remaining majority to low bits.
Concretely, HyQuant consists of three components:
\textbf{(i) vertical-line-aware high-precision retention},
\textbf{(ii) Prefill-stage hybrid-precision quantized attention},
and \textbf{(iii) Decode-stage hybrid low-bit KV cache with fused attention}.

\begin{figure}[t]
  \centering
  \includegraphics[width=1\linewidth,angle=0]{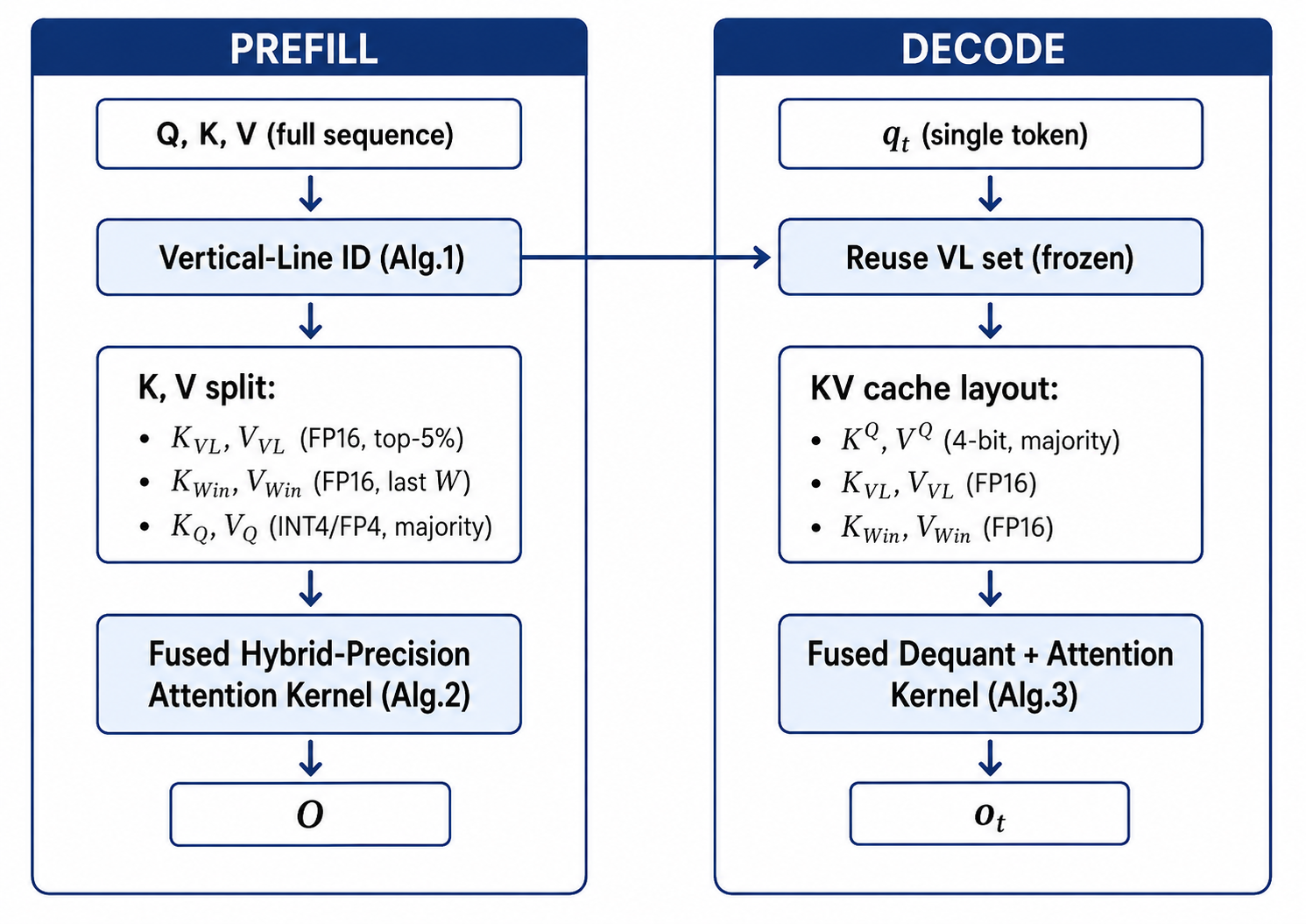}
  \caption{\textbf{Overview of HyQuant.}
  HyQuant retains vertical-line positions and a local window in full precision, while quantizing the remaining majority for efficient prefill and decode.}
  \label{fig:hyquant_flowchart}
\end{figure}

\subsection{Vertical-line Awareness}
\label{sec:method:vertical_awareness}

\begin{figure}[t]
  \centering
  \includegraphics[width=0.9\linewidth]{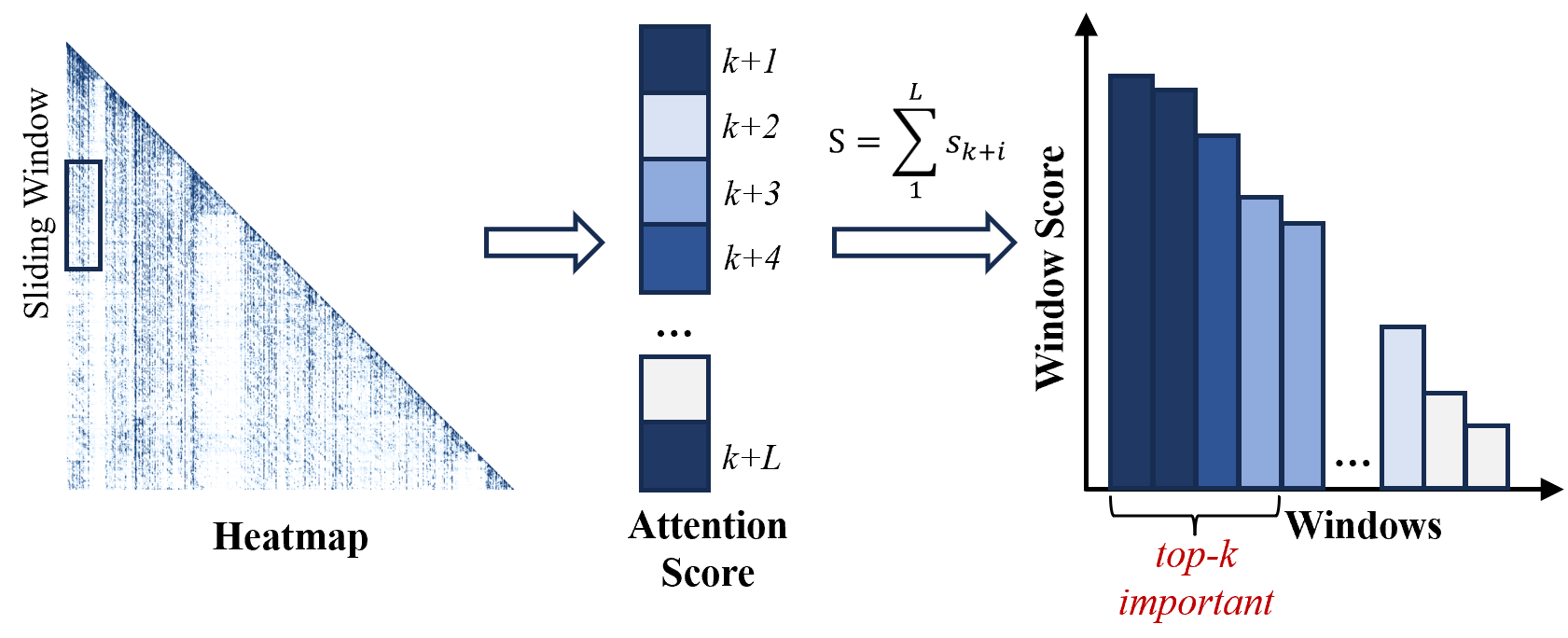}
  \caption{\textbf{Vertical-line awareness process.}
  HyQuant estimates column-wise attention mass to identify persistent vertical-line positions for full-precision retention.}
  \label{fig:vertical_awareness_process}
\end{figure}

\paragraph{Token sets.}
For each layer/head, we partition the key positions into three disjoint subsets:
\begin{equation}
\mathcal{K} = \mathcal{K}_{\text{VL}} \ \cup\  \mathcal{K}_{\text{Win}} \ \cup\  \mathcal{K}_{\text{Q}},
\end{equation}
where $\mathcal{K}_{\text{VL}}$ denotes the \emph{vertical-line} positions, $\mathcal{K}_{\text{Win}}$ denotes a fixed sliding window in the end of prefill query, and $\mathcal{K}_{\text{Q}}$ denotes the remaining majority to be quantized.
Vertical-line positions are selected from the non-window prefix, making the three subsets disjoint by construction.

\paragraph{Window definition.}
Given the current query index $t$, the window keys are
\begin{equation}
\mathcal{K}_{\mathrm{Win}}=\{k\mid k\in[\max(0,\,t_0-W),\;t_0-1]\}
\label{eq:win_def}
\end{equation}
where $W$ is a pre-specified window size.
We denote the non-window prefix by
\begin{equation}
\mathcal{K}_{\text{pre}}(t) = \mathcal{K} \setminus \mathcal{K}_{\text{Win}}(t).
\label{eq:pre_def}
\end{equation}

\paragraph{Vertical-line identification.}
We maintain a running importance score for each non-window key position that measures the accumulated \emph{column mass} (vertical attention mass).
Let $a_{t,k}$ be the attention probability from query $t$ to key $k$.
We define the vertical-line score as
\begin{equation}
S(k) = \sum_{t \in \mathcal{T}} a_{t,k},
\label{eq:vl_score}
\end{equation}
where $\mathcal{T}$ is the set of queries considered (e.g., within the current segment or within a running buffer).
We then select the top-$\rho$ fraction from the non-window prefix as vertical-line positions:
\begin{equation}
\mathcal{K}_{\text{VL}}
=
\mathrm{TopK}\big(S|_{\mathcal{K}_{\text{pre}}},\ \rho \cdot |\mathcal{K}_{\text{pre}}|\big),
\qquad \rho \ll 1.
\label{eq:vl_topk}
\end{equation}
In practice, $\rho$ is small.
Importantly, $S(k)$ can be updated with negligible overhead using a lightweight reduction, and the window size $W$ is fixed in advance; therefore, \textbf{vertical-line awareness introduces virtually no extra preprocessing cost} compared to sparsification pipelines.
The detailed algorithm is shown in Algorithm~\ref{alg:vertical-indices}.

\subsection{Prefill-stage Hybrid-Precision Quantized Attention}
\label{sec:method:prefill}

\paragraph{Hybrid-precision quantized attention in Prefill.}
In Prefill, attention is compute-intensive and dominated by large GEMMs.
HyQuant quantizes the majority of attention computation, while retaining full precision on $\mathcal{K}_{\text{VL}}$ and $\mathcal{K}_{\text{Win}}$.

Specifically, for a query block $Q$, we conceptually split keys/values into
\begin{equation}
(K,V) = (K_{\text{Q}}, V_{\text{Q}}) \ \cup\ (K_{\text{FP}}, V_{\text{FP}}),
\end{equation}
where $(K_{\text{Q}},V_{\text{Q}})$ correspond to $\mathcal{K}_{\text{Q}}$, and $(K_{\text{FP}},V_{\text{FP}})$ correspond to $\mathcal{K}_{\text{VL}} \cup \mathcal{K}_{\text{Win}}$.

We compute attention with a fused operator:
\begin{equation}
O =
\mathrm{Softmax}\!\left(
\frac{Q\hat{K}_{\text{Q}}^\top}{\sqrt{d}}
\ \Vert\
\frac{QK_{\text{FP}}^\top}{\sqrt{d}}
\right)
\cdot
\left(\hat{V}_{\text{Q}} \ \Vert\ V_{\text{FP}}\right),
\label{eq:prefill_hybrid}
\end{equation}
where $\hat{K}_{\text{Q}}, \hat{V}_{\text{Q}}$ are dequantized values of low-bit $(K_{\text{Q}},V_{\text{Q}})$, and $\Vert$ denotes concatenation along the key dimension.
The concatenation order follows the physical KV layout used by the fused kernel.
Crucially, we \emph{fuse} the full-precision path and the quantized path into a single kernel so that the operator structure remains FlashAttention-like (online softmax with blockwise scanning), and the additional overhead over naive low-bit attention quantization is minimal.

\paragraph{Implementation note.}
In our implementation, we use lower precision for the bulk computation (e.g., INT8/FP8, INT4/FP4 depending on the backend), while retaining the vertical-line and window tiles in FP16/BF16.
This hybrid-precision execution preserves the accuracy-critical attention mass with a tiny full-precision budget.
The detailed algorithm is shown in Algorithm~\ref{alg:hyquant_prefill}.

\subsection{Decode-stage Low-bit KV Cache and Fused Attention}
\label{sec:method:decode}

\paragraph{Low-bit KV cache with selective full precision.}
Decode is memory-intensive: each step reads a large prefix KV.
HyQuant stores the KV cache in a hybrid-precision manner:
\begin{equation}
(K_{1:t},V_{1:t}) =
(K^{\text{Q}}_{1:t},V^{\text{Q}}_{1:t}) \ \cup\ (K^{\text{FP}}_{1:t},V^{\text{FP}}_{1:t}),
\end{equation}
where full precision is reserved for keys in $\mathcal{K}_{\text{VL}} \cup \mathcal{K}_{\text{Win}}(t)$, and the remaining majority is stored in low bits.
This directly reduces KV memory footprint and bandwidth pressure.

\paragraph{Fused dequantization and attention.}
To avoid extra memory traffic, we fuse dequantization into the decode attention kernel:
when scanning a quantized KV block, we dequantize on the fly and immediately apply the values to score and value accumulation using FlashAttention-style online softmax.
Formally, for a quantized block $(K^{\text{Q}}_b,V^{\text{Q}}_b)$,
\begin{equation}
\hat{K}_b = \mathrm{DeQuant}(K^{\text{Q}}_b), 
\hat{V}_b = \mathrm{DeQuant}(V^{\text{Q}}_b),
\end{equation}
and the kernel updates the running softmax state and output accumulator using $(\hat{K}_b,\hat{V}_b)$ without materializing full-precision intermediates.
For blocks belonging to $\mathcal{K}_{\text{VL}} \cup \mathcal{K}_{\text{Win}}(t)$, we directly load full-precision keys/values.

\paragraph{Summary.}
By retaining only a tiny set of vertical-line positions and a local window in full precision, HyQuant controls the dominant quantization errors, while quantizing the remaining majority to achieve low-bit efficiency.
In Prefill, HyQuant preserves low-bit attention efficiency while reducing numerical error; in Decode, it reduces KV memory traffic together with fused dequantization for bandwidth efficiency.
The detailed algorithm is shown in Algorithm~\ref{alg:hyquant_decode}.
\section{Experiments}
\label{sec:experiments}

\subsection{Experimental Setup}
\label{sec:exp_setup}

\paragraph{Models.}
We primarily evaluate HyQuant on \textbf{Qwen3-8B}~\cite{yang2025qwen3technicalreport} and further validate its generality on \textbf{Llama-3.1-8B-Instruct}~\cite{grattafiori2024llama3herdmodels}, \textbf{GLM-4-9B-0414}~\cite{glm2024chatglm}, and \textbf{Qwen3-32B}~\cite{yang2025qwen3technicalreport}.
We focus on long-context reasoning capability and adopt predominantly chain-of-thought task settings to cover multi-step reasoning and information aggregation under long sequences.

\paragraph{Baselines.}
We compare against representative full-precision and low-bit inference baselines:
\textbf{(i) FlashAttention-2 (FA2)}~\cite{dao2023flashattention2}, which serves as the full-precision accuracy and latency baseline;
\textbf{(ii) KIVI}~\cite{liu2024kivi}, a KV-cache quantization baseline for reducing memory and bandwidth overhead during long-context inference;
\textbf{(iii) KVTuner}~\cite{li2025kvtuner}, a sensitivity-aware mixed-precision KV-cache quantization baseline;
and \textbf{(iv) SageAttention}~\cite{zhang2025sageattentionaccurate8bitattention}, an attention quantization baseline that mitigates activation outliers and uses a numerically stable computation path.
For each model, we report the applicable baselines depending on implementation availability. Detailed settings can be seen in \ref{app:exp_config}.

\paragraph{Implementation Details (HyQuant).}
Unless specified otherwise, all experiments are conducted on an \textbf{NVIDIA H100} GPU.
HyQuant adopts a hybrid-precision design for both Prefill attention and Decode KV-cache compression.
We retain two categories of positions in full precision: (i) the online-identified top-$5\%$ vertical-line tokens, and (ii) tokens inside a local sliding window.
All remaining KV positions are stored in Key-4bit, Value-4bit quantized formats.
Quantized caches are recovered through dequantization and hybrid-precision computation to balance accuracy and efficiency.

\paragraph{Benchmarks and metrics.}
For accuracy, we evaluate mathematical reasoning on \textbf{GSM8K}~\cite{cobbe2021trainingverifierssolvemath} and \textbf{MATH500}~\cite{hendrycks2021math,lightman2023lets}, and assess long-context capability on \textbf{LongBench v1}~\cite{bai2024longbench}.
For efficiency and operator-level analysis, we focus on the Prefill and Decode stages and report:
(i) \textbf{Prefill operator-level error}: using full-precision FlashAttention-2 as the reference, we compute the layer-wise MSE of the intermediate attention output
(i.e., the input to $o_{\text{proj}}$) to quantify numerical deviations introduced by low-bit computation, and compare against SageAttention;
(ii) \textbf{Decode latency}: under the same decoding setup, we report the per-step total decoding latency (ms/token) across different prefix lengths and compare with FA2, together with the relative speedup.
All latency numbers are collected after sufficient warmup and averaged over multiple runs.

\begin{table*}[t]
  \centering
  \scriptsize
  \setlength{\tabcolsep}{2.2pt}
  \renewcommand{\arraystretch}{1.08}
  \caption{\textbf{LongBench v1 results on Qwen3-8B (thinking mode).}
  We compare full-precision attention (FA2), KIVI, KVTuner, SageAttention, and HyQuant (Ours).
  \textbf{Avg.} denotes the arithmetic mean over the 11 evaluated LongBench tasks reported in this table.}
  \label{tab:longbenchv1_main_qwen}
  \resizebox{\textwidth}{!}{%
  \begin{tabular}{l*{12}{c}}
    \toprule
    \textbf{Method} &
    \multicolumn{3}{c}{\textbf{Single-Document QA}} &
    \multicolumn{3}{c}{\textbf{Multi-Document QA}} &
    \multicolumn{2}{c}{\textbf{Summarization}} &
    \multicolumn{2}{c}{\textbf{Few-shot Learning}} &
    \multicolumn{1}{c}{\textbf{Semantic}} &
    \textbf{Avg.} \\
    \cmidrule(lr){2-4}\cmidrule(lr){5-7}\cmidrule(lr){8-9}
    \cmidrule(lr){10-11}\cmidrule(lr){12-12}
    &
    \textbf{NarrQA} & \textbf{Qasper} & \textbf{MF-en} &
    \textbf{HotpotQA} & \textbf{2Wiki} & \textbf{Musique} &
    \textbf{QMSum} & \textbf{MNews} &
    \textbf{Trivia} & \textbf{SAMSum} & \textbf{Pretrieve-en} &
    \\
    \midrule
    FA2 (full-precision)             & 30.37 & 41.68 & \textbf{50.34} & \textbf{50.42} & 39.69 & 33.88 & 13.11 & 10.85 & \textbf{97.50} & \textbf{33.80} & \textbf{88.81} & 44.59 \\
    KIVI (K4V4)                      & 24.76 & 35.13 & 45.51 & 42.55 & 31.04 & 18.92 & 10.51 & 8.19 & 88.87 & 27.56 & 81.41 & 37.68 \\
    SageAttention                    & 19.45 & 34.54 & 41.95 & 46.59 & 44.33 & 24.31 & 18.67 & \textbf{23.23} & 74.93 & 32.62 & 58.86 & 38.13 \\
    KVTuner (4bit)                   & 13.74 & 32.45 & 45.32 & 37.68 & \textbf{54.18} & 31.92 & \textbf{20.20} & 10.67 & 94.70 & 29.32 & 74.81 & 40.45 \\
    \textbf{HyQuant (K4V4, top-5\%)} & \textbf{32.28} & \textbf{41.82} & 49.74 & 49.75 & 43.31 & \textbf{35.24} & 15.39 & 11.94 & 96.65 & 33.05 & 86.32 & \textbf{45.04} \\
    \bottomrule
  \end{tabular}}
\end{table*}

\begin{table*}[t]
  \centering
  \scriptsize
  \setlength{\tabcolsep}{2.2pt}
  \renewcommand{\arraystretch}{1.08}
  \caption{\textbf{LongBench v1 results on Llama-3.1-8B-Instruct.}
  We compare full-precision attention (FA2), KVTuner, SageAttention, and HyQuant (Ours).
  \textbf{Avg.} denotes the reported average score over the evaluated LongBench tasks.}
  \label{tab:longbenchv1_main_llama}

  \resizebox{\textwidth}{!}{%
  \begin{tabular}{l*{12}{c}}
    \toprule
    \textbf{Method} &
    \multicolumn{3}{c}{\textbf{Single-Document QA}} &
    \multicolumn{3}{c}{\textbf{Multi-Document QA}} &
    \multicolumn{2}{c}{\textbf{Summarization}} &
    \multicolumn{2}{c}{\textbf{Few-shot Learning}} &
    \multicolumn{1}{c}{\textbf{Semantic}} &
    \textbf{Avg.} \\
    \cmidrule(lr){2-4}\cmidrule(lr){5-7}\cmidrule(lr){8-9}
    \cmidrule(lr){10-11}\cmidrule(lr){12-12}
    &
    \textbf{NarrQA} & \textbf{Qasper} & \textbf{MF-en} &
    \textbf{HotpotQA} & \textbf{2Wiki} & \textbf{Musique} &
    \textbf{QMSum} & \textbf{MNews} &
    \textbf{Trivia} & \textbf{SAMSum} & \textbf{Pretrieve-en} &
    \\
    \midrule
    FA2 (full-precision)             & 30.96 & 45.16 & 54.60 & \textbf{55.63} & 45.70 & 31.69 & 24.26 & 26.17 & 88.36 & 12.48 & 97.90 & 46.63 \\
    SageAttention                    & 31.16 & 43.29 & 55.01 & 51.81 & \textbf{46.97} & 30.99 & \textbf{24.96} & 26.64 & 86.67 & 6.87 & 98.15 & 45.68 \\
    KVTuner (4bit)                   & \textbf{31.29} & 42.06 & 53.97 & 50.11 & 45.57 & \textbf{32.49} & 24.91 & 27.08 & \textbf{88.42} & 7.68 & 98.70 & 45.66 \\
    \textbf{HyQuant (K4V4, top-5\%)} & 30.75 & \textbf{46.51} & \textbf{55.84} & 51.22 & 45.30 & 31.90 & 24.84 & \textbf{27.21} & 87.67 & \textbf{13.75} & \textbf{99.12} & \textbf{46.73} \\
    \bottomrule
  \end{tabular}}
\end{table*}

\begin{table*}[t!]
  \centering
  \scriptsize
  \setlength{\tabcolsep}{2.2pt}
  \renewcommand{\arraystretch}{1.08}
  \caption{\textbf{LongBench v1 results on GLM-4-9B-0414.}
  We compare full-precision attention (FA2), SageAttention, KVTuner, and HyQuant (Ours).
  \textbf{Avg.} denotes the reported average score over the evaluated LongBench tasks.}
  \label{tab:longbenchv1_main_glm4}

  \resizebox{\textwidth}{!}{%
  \begin{tabular}{l*{12}{c}}
    \toprule
    \textbf{Method} &
    \multicolumn{3}{c}{\textbf{Single-Document QA}} &
    \multicolumn{3}{c}{\textbf{Multi-Document QA}} &
    \multicolumn{2}{c}{\textbf{Summarization}} &
    \multicolumn{2}{c}{\textbf{Few-shot Learning}} &
    \multicolumn{1}{c}{\textbf{Semantic}} &
    \textbf{Avg.} \\
    \cmidrule(lr){2-4}\cmidrule(lr){5-7}\cmidrule(lr){8-9}
    \cmidrule(lr){10-11}\cmidrule(lr){12-12}
    &
    \textbf{NarrQA} & \textbf{Qasper} & \textbf{MF-en} &
    \textbf{HotpotQA} & \textbf{2Wiki} & \textbf{Musique} &
    \textbf{QMSum} & \textbf{MNews} &
    \textbf{Trivia} & \textbf{SAMSum} & \textbf{Pretrieve-en} &
    \\
    \midrule
    FA2 (full-precision)             & 24.02 & 39.67 & 51.49 & 40.08 & \textbf{50.86} & 26.43 & 22.06 & 23.86 & 85.73 & \textbf{37.72} & 91.26 & 44.83 \\
    SageAttention                    & 24.85 & \textbf{43.30} & \textbf{52.05} & 47.10 & 47.66 & 29.42 & \textbf{23.16} & 21.37 & \textbf{88.46} & 35.64 & 90.25 & 45.75 \\
    KVTuner (4bit)                   & 23.51 & 41.39 & 50.14 & 46.65 & 46.57 & \textbf{32.16} & 23.05 & 24.25 & 85.92 & 37.14 & 91.15 & 45.63 \\
    \textbf{HyQuant (K4V4, top-5\%)} & \textbf{25.41} & 40.07 & 49.97 & \textbf{48.70} & 49.18 & 26.30 & 22.42 & \textbf{24.52} & 87.46 & 37.48 & \textbf{92.10} & \textbf{45.78} \\
    \bottomrule
  \end{tabular}}
\end{table*}

\begin{table*}[t]
  \centering
  \scriptsize
  \setlength{\tabcolsep}{2.2pt}
  \renewcommand{\arraystretch}{1.08}
  \caption{\textbf{LongBench v1 results on Qwen3-32B (thinking mode).}
  We compare full-precision attention (FA2), KIVI, KVTuner, and HyQuant (Ours).
  \textbf{Avg.} denotes the reported average score over the evaluated LongBench tasks.}
  \label{tab:longbenchv1_main_qwen_32B}

  \resizebox{\textwidth}{!}{%
  \begin{tabular}{l*{12}{c}}
    \toprule
    \textbf{Method} &
    \multicolumn{3}{c}{\textbf{Single-Document QA}} &
    \multicolumn{3}{c}{\textbf{Multi-Document QA}} &
    \multicolumn{2}{c}{\textbf{Summarization}} &
    \multicolumn{2}{c}{\textbf{Few-shot Learning}} &
    \multicolumn{1}{c}{\textbf{Semantic}} &
    \textbf{Avg.} \\
    \cmidrule(lr){2-4}\cmidrule(lr){5-7}\cmidrule(lr){8-9}
    \cmidrule(lr){10-11}\cmidrule(lr){12-12}
    &
    \textbf{NarrQA} & \textbf{Qasper} & \textbf{MF-en} &
    \textbf{HotpotQA} & \textbf{2Wiki} & \textbf{Musique} &
    \textbf{QMSum} & \textbf{MNews} &
    \textbf{Trivia} & \textbf{SAMSum} & \textbf{Pretrieve-en} &
    \\
    \midrule
    FA2 (full-precision)             & 34.93 & \textbf{46.38} & 43.79 & \textbf{54.18} & 50.57 & \textbf{45.59} & 17.11 & 14.74 & \textbf{90.60} & 36.80 & \textbf{100.00} & \textbf{48.61} \\
    KIVI                             & 19.89 & 42.21 & 37.87 & 47.42 & 46.16 & 41.30 & 12.67 & 11.16 & 80.37 & 28.68 & 94.00 & 41.98 \\
    KVTuner (4bit)                   & 32.97 & 45.64 & 44.89 & 52.13 & 50.47 & 45.05 & 16.79 & 14.82 & \textbf{90.60} & \textbf{37.10} & \textbf{100.00} & 48.22 \\
    \textbf{HyQuant (K4V4, top-5\%)} & \textbf{45.56} & 41.05 & \textbf{46.17} & 44.44 & \textbf{53.12} & 43.76 & \textbf{17.15} & \textbf{14.89} & \textbf{90.60} & 36.35 & \textbf{100.00} & 48.46 \\
    \bottomrule
  \end{tabular}}
\end{table*}

\subsection{Main Results}
\label{sec:main_results}

\subsubsection{Benchmark Results}
Tables~\ref{tab:longbenchv1_main_qwen}, \ref{tab:longbenchv1_main_llama}, \ref{tab:longbenchv1_main_glm4}, \ref{tab:longbenchv1_main_qwen_32B}, and~\ref{tab:math_main_qwen} summarize the end-to-end benchmark results.
Across Qwen3-8B, Llama-3.1-8B-Instruct, GLM-4-9B-0414, and Qwen3-32B, HyQuant generally preserves the performance of full-precision attention (FA2) and improves over the applicable strict low-bit baselines, including KIVI, KVTuner, and SageAttention, in most settings.
This indicates that a hybrid-precision strategy that \emph{retains a small set of vertical-line tokens together with a local window in full precision} is effective for preserving long-context understanding and mathematical reasoning ability.
For Qwen3-8B, we enable thinking mode in end-to-end evaluation to better reflect realistic long-CoT generation and KV reuse.

\subsubsection{Operator-level evaluation: Prefill error and Decode latency}
\label{sec:op_prefill_decode}

\begin{figure}[t]
  \centering
  \includegraphics[width=0.9\linewidth]{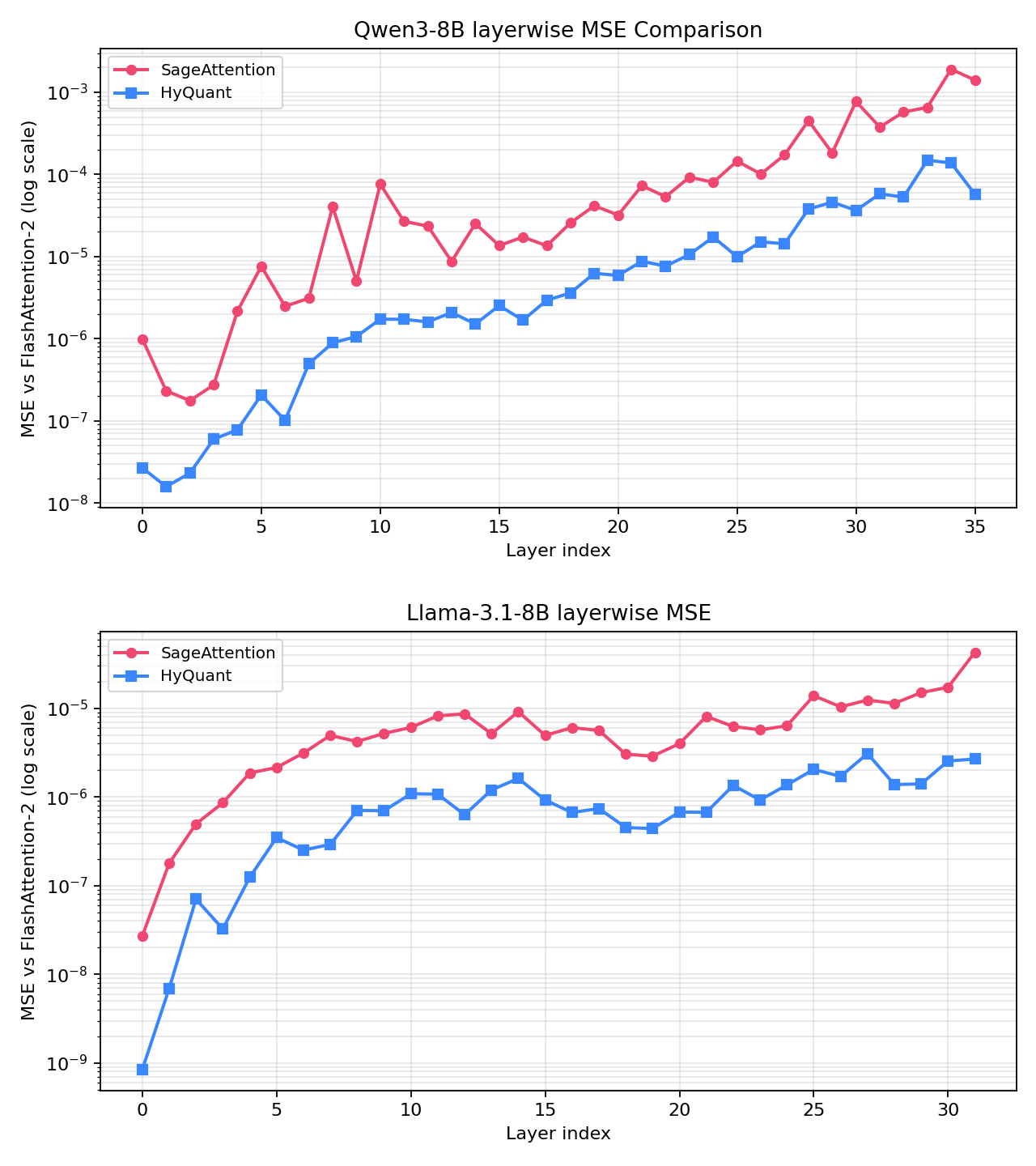}
  \caption{\textbf{Layer-wise MSE reduction factor in Prefill.}
  Using FA2 as the reference, we compute the MSE of the intermediate attention output (input to $o_{\text{proj}}$) per layer.
  We plot the MSE reduction factor $\mathrm{MSE}_{\text{Sage}}/\mathrm{MSE}_{\text{HyQuant}}$, where larger is better.}
  \label{fig:prefill_mse_layer}
\end{figure}

\paragraph{Prefill: layer-wise MSE.}
Since the Prefill latency of HyQuant is comparable to that of SageAttention under our implementation, we focus on numerical deviations rather than additional prefill speedup claims in this stage.
We use full-precision FA2 as the reference and compute the \textbf{MSE} of the intermediate attention output
(i.e., the input to $o_{\text{proj}}$) for each layer.
We further report the \textbf{MSE reduction factor} of HyQuant over SageAttention, defined as
$\mathrm{MSE}_{\text{Sage}} / \mathrm{MSE}_{\text{HyQuant}}$.
As shown in Fig.~\ref{fig:prefill_mse_layer}, we visualize the first five layers as representative examples.
HyQuant achieves a large reduction in MSE over SageAttention in these layers, demonstrating that
\emph{retaining vertical-line tokens and the local window in full precision} effectively suppresses error amplification under low-bit computation.

\paragraph{Decode: decoding latency vs.\ FA2.}
In the Decode stage, we compare HyQuant against FA2 in terms of decode attention-kernel latency under different prefix lengths.
Table~\ref{tab:decode_latency_vs_len} summarizes the kernel-level results: as the context grows, HyQuant achieves increasingly larger gains, reaching about $3.58\times$ speedup over FA2 at the 32K prefix length, while the corresponding end-to-end decode speedup is more moderate.
This highlights the benefits of hybrid-precision KV storage and fused dequantization for bandwidth- and memory-bound attention kernels in long-context decoding.

\subsection{High parallel setting speedup}
While HyQuant achieves significant speedup under batch size 1 and maintains accuracy, its performance under high parallel, reality settings remains unknown. To demonstrate its ability, we compare HyQuant with Flashattention-2~\cite{dao2023flashattention2}, KIVI~\cite{liu2024kivi}, KVTuner~\cite{li2025kvtuner} under multi-batch settings. Results can be seen in \ref{tab:throughput}.

\begin{table}[t]
\caption{Throughput (tokens/s) under increasing batch size on Qwen3-8B with a 32K prefix, single H100-80GB. All methods except FA2 are tested under 4-bit quantization for KV cache.}
\centering
\small
\begin{tabular}{lcccc}
\toprule
Batch & FA2 & HyQuant & KIVI & KVTuner \\
\midrule
1  & 42.8  & \textbf{43.7}  & 23.7 & 24.2 \\
4  & \textbf{122.2} & 84.2  & 32.5 & 33.4 \\
8  & 152.7 & \textbf{155.6} & 34.9 & 35.6 \\
16 & OOM   & \textbf{231.6} & OOM  & OOM  \\
32 & OOM   & OOM            & OOM  & OOM  \\
\bottomrule
\end{tabular}
\label{tab:throughput}
\end{table}
The results show that HyQuant is the only method that operates in batch size 16, showing its superiority, while outperforms other methods in terms of speed in lower batch settings. This is a mutual effect from KV cache compression and employing a fused kernel, decreasing bandwidth requirements and peak memory occupation. Compared with our implementations of KIVI~\cite{liu2024kivi} and KVTuner~\cite{li2025kvtuner}, which follow a ``dequantize-then-attend'' strategy---the quantized KV cache is dequantized before running standard attention---HyQuant adopts a fused ``dequantize-in-attention'' design that unpacks the KV cache within a single online-softmax pass and never materializes the full-precision cache. The results demonstrate our method's advantage, which employs an operator-level co-design of compression and computation.

\paragraph{Summary.}
In our experiments on Qwen3-8B, Llama-3.1-8B-Instruct, GLM-4-9B-0414, and Qwen3-32B, HyQuant achieves a favorable accuracy--efficiency trade-off with strong decoding efficiency.
Across the evaluated benchmarks, HyQuant remains close to the full-precision baseline while avoiding the larger degradation observed in strict low-bit quantization baselines.
On some datasets, HyQuant slightly exceeds FA2, which we treat as normal evaluation variance rather than evidence that quantization improves the underlying model capability.

\begin{table}[t!]
  \centering
  \small
  \setlength{\tabcolsep}{4.5pt}
  \renewcommand{\arraystretch}{1.15}
  \caption{\textbf{Decode kernel latency under different prefix lengths (ms/token).}
  Speedup is computed as FA2 / HyQuant.}
  \label{tab:decode_latency_vs_len}

  \begin{tabularx}{\columnwidth}{>{\centering\arraybackslash}X
                                >{\centering\arraybackslash}X
                                >{\centering\arraybackslash}X
                                >{\centering\arraybackslash}X}
    \toprule
    \textbf{Sequence length} & \textbf{FA2} & \textbf{HyQuant} & \textbf{Speedup} \\
    \midrule
     1{,}024  & 0.224 & 0.170 & 1.32$\times$ \\
     2{,}048  & 0.416 & 0.173 & 2.40$\times$ \\
     4{,}096  & 0.805 & 0.263 & 3.06$\times$ \\
     8{,}192  & 1.604 & 0.478 & 3.36$\times$ \\
    16{,}384  & 3.181 & 0.903 & 3.52$\times$ \\
    32{,}768  & 6.354 & 1.775 & 3.58$\times$ \\
    \bottomrule
  \end{tabularx}
\end{table}

\begin{table}[t!]
  \centering
  \small
  \setlength{\tabcolsep}{4.5pt}
  \renewcommand{\arraystretch}{1.15}
  \caption{\textbf{End-to-end decode speed relative to FA2 under different prefix lengths.}
  Values are normalized by the FA2 baseline; values above $1.0\times$ indicate speedup.}
  \label{tab:decode_latency_vs_len_endtoend}

  \begin{tabularx}{\columnwidth}{>{\centering\arraybackslash}X
                                >{\centering\arraybackslash}X
                                >{\centering\arraybackslash}X
                                >{\centering\arraybackslash}X}
    \toprule
    \textbf{Sequence length} & \textbf{KIVI} & \textbf{KVTuner} & \textbf{HyQuant} \\
    \midrule
     1{,}024  & 0.72$\times$ & 0.78$\times$ & 1.04$\times$ \\
     2{,}048  & 0.72$\times$ & 0.79$\times$ & 1.07$\times$ \\
     4{,}096  & 0.73$\times$ & 0.80$\times$ & 1.09$\times$ \\
     8{,}192  & 0.73$\times$ & 0.80$\times$ & 1.12$\times$ \\
    16{,}384  & 0.71$\times$ & 0.74$\times$ & 1.13$\times$ \\
    32{,}768  & 0.69$\times$ & 0.72$\times$ & 1.17$\times$ \\
    \bottomrule
  \end{tabularx}
\end{table}

\subsection{Ablations}
\label{sec:ablation}

\subsubsection{Importance of the sliding window and vertical-line awareness}
To validate the effectiveness of the sliding window and vertical-line awareness design, we conduct ablation studies on Qwen3-8B with LongBench v1.
The results show that both the local full-precision window and vertical-line-aware retention reduce the MSE during the Prefill stage.

\subsubsection{Vertical-line token ratio (\texorpdfstring{$k$}{k})}
We vary the retained vertical-line token ratio $k$ while keeping other settings fixed to study the accuracy--efficiency trade-off.
As shown in Table~\ref{tab:ablation_k}, increasing the top-$k$ ratio generally improves accuracy and reduces quantization error, but also increases the full-precision KV budget.
We therefore use top-$5\%$ as a practical trade-off unless otherwise specified.

\subsubsection{Local full-precision window size}
We vary the local full-precision window size to evaluate sensitivity to local-context dependency and verify its complementarity with vertical-line-aware retention.
As shown in Table~\ref{tab:ablation_win}, a larger window size slightly improves accuracy, likely because more recent tokens remain in full precision during inference.

\subsection{Additional Cost Overhead}
\label{sec:additional_overhead}

The additional runtime overhead mainly comes from vertical-line identification, including accumulating query vectors.
Our evaluation shows that this additional vertical-line identification overhead accounts for only \textbf{3\% to 5\%} of the total runtime.
This minor overhead is largely offset by the hybrid-precision attention kernels and Triton implementations.

For memory overhead, HyQuant buffers at most 64 query vectors in FP16, resulting in an auxiliary memory cost of $64 \times H_Q \times d \times 2$ bytes.
Keeping 5\% of vertical-line tokens in full precision increases the non-window KV cache size by about 15\% compared with strict 4-bit quantization; the total overhead additionally depends on the local-window size.
In practice, the improvements in end-to-end speed and accuracy offset this additional memory overhead. Detailed analysis can be seen in \ref{app:memory_overhead}.
\section{Conclusion}
In this paper, we present HyQuant, a hybrid-precision quantization framework for efficient long-context LLM inference.
HyQuant identifies vertical-line tokens using accumulated column-wise attention scores, retains these tokens and a local window in full precision, and quantizes the remaining majority to low-bit formats.
We further implement fused operators that integrate low-bit computation, full-precision retention, and on-the-fly dequantization for Prefill and Decode execution.
Experimental results show that HyQuant achieves 1.32$\times$ to 3.58$\times$ decode-kernel speedup and 1.04$\times$ to 1.17$\times$ end-to-end decode speedup while maintaining near-full-precision accuracy across multiple long-context and reasoning benchmarks.

\section*{Acknowledgments}
This work is sponsored by National Natural Science Foundation of China (No.62602410), Natural Science Foundation of Shanghai (No.24ZR1430600), Tencent Basic Platform Technology Rhino-Bird Focused Research Program and Shanghai Key Laboratory of Trusted Data Circulation and Governance, and Web3.

\section*{Limitations}
\begin{itemize}
    \item The effect of retaining critical vertical-line tokens is most visible in long-context tasks. For short-context tasks, HyQuant may not provide the same level of improvement, but it still maintains competitive performance relative to the full-precision baseline. See Appendix~\ref{app:short_context}, Table~\ref{tab:short_context_qwen}.
    \item Our experiments are conducted on NVIDIA H100 GPUs. On high-end GPUs, end-to-end speedups can be less pronounced at short context lengths because the decoding pipeline is not fully memory-bound. Due to GPU memory limitations, we have not evaluated larger-scale models beyond Qwen3-32B, such as Qwen3-80B.
    \item  We have made a broad assessment across tasks and models. However, it is unknown whether our method remains effective in agent / coding-related settings.
\end{itemize}

\FloatBarrier

\bibliography{custom}


\clearpage
\appendix
\raggedbottom  

\section{Implementation Details and Algorithms}
\label{app:algorithms}

This appendix provides implementation-level details for HyQuant.
We include the algorithmic steps that are omitted from the main text due to
space constraints. Algorithm~\ref{alg:vertical-indices} describes how HyQuant
identifies vertical-line tokens from accumulated attention signals.
Algorithms~\ref{alg:hyquant_prefill} and~\ref{alg:hyquant_decode} describe the
hybrid-precision attention operators used in the prefill and decode stages.
Algorithm~\ref{alg:hyquant_freeze} gives the cache reorganization procedure at
the prefill-to-decode boundary. In the algorithms below, \VTI{} abbreviates
\textsc{VerticalTokenIdentification}.


Algorithm~\ref{alg:vertical-indices} implements the lightweight vertical-line
token selection used by HyQuant. The algorithm first excludes the recent local
window and then estimates the column-wise importance of the non-window prefix.
For GQA models, it computes the scores in a GQA-native form without explicitly
materializing repeated KV heads. The returned indices are used as the
full-precision vertical-line token set in both prefill and decode.

\subsection{Vertical-Line Token Identification}
\label{app:vertical_identification}

\begin{algorithm}[H]
\caption{Vertical Token Identification}
\label{alg:vertical-indices}
\footnotesize
\begin{algorithmic}[1]
\Require Query $Q \in \mathbb{R}^{B \times H_Q \times L_Q \times D}$
\Require Key $K \in \mathbb{R}^{B \times H \times L_K \times D}$, where $H \in \{H_Q,H_{KV}\}$
\Require Window size $W$, ratio $r \in (0,1]$, GQA factor $g=H_Q/H_{KV}$
\Ensure Vertical key indices $\mathcal{I} \in \mathbb{Z}^{B \times H_Q \times k}$

\State $P \gets \max(0, L_K-W)$ \Comment{exclude the local window}
\If{$P=0$}
    \State \Return $\varnothing$
\EndIf

\State $t \gets \min(W,L_Q)$
\State $\bar{Q} \gets \frac{1}{t}\sum_{i=L_Q-t+1}^{L_Q} Q_{:,:,i,:}$
\Comment{tail-query proxy}
\State $K^{\mathrm{pre}} \gets K_{:,:,1:P,:}$

\If{$H=H_Q$}
    \State $S \gets \bar{Q}{K^{\mathrm{pre}}}^{\top}$
\Else
    \State Reshape $\bar{Q} \to \bar{Q}^{g} \in \mathbb{R}^{B \times H_{KV} \times g \times D}$
    \State $S \gets \texttt{einsum}(\text{``}bhrd,bhpd\!\to\! bhrp\text{''},\bar{Q}^{g},K^{\mathrm{pre}})$
    \State Reshape $S \to \mathbb{R}^{B \times H_Q \times P}$
\EndIf

\State $k \gets \min(P,\max(1,\lceil P\cdot r\rceil))$
\State $\mathcal{I} \gets \textsc{TopK}(S,k,\mathrm{dim}{=}{-1},\mathrm{largest})$
\State \Return $\mathcal{I}$ \Comment{indices in prefix $[0,P)$}
\end{algorithmic}
\end{algorithm}


\subsection{Hybrid-Precision Prefill Operator}
\label{app:prefill_algorithm}

Algorithm~\ref{alg:hyquant_prefill} shows the prefill-stage hybrid-precision
attention operator. The key/value sequence is partitioned into three regions:
a low-bit quantizable prefix, a full-precision vertical-line region, and a
full-precision local window. HyQuant scans these regions in a segmented
FlashAttention-style online softmax, so the quantized and full-precision paths
are fused into a single logical operator rather than executed as separate
attention calls.

\begin{algorithm}[H]
\caption{\textbf{HyQuant Prefill}: Hybrid-Precision Quantized Attention}
\label{alg:hyquant_prefill}
\footnotesize
\begin{algorithmic}[1]
\Require Query block $Q_b$; full $K,V$; window size $W$; vertical ratio $\rho$
\Require Bitwidth $B \in \{\mathrm{INT8},\mathrm{FP8}\}$; tile size $B_N$; scale $\alpha=1/\sqrt{d}$
\Ensure Output block $O_b$

\Statex \textcolor{vlblue}{\# Stage 1: identify vertical-line keys and reorder KV}
\State $\mathcal{K}_{\mathrm{Win}} \gets$ last $W$ keys
\State $\mathcal{I}_{\mathrm{VL}} \gets \VTI(Q,K,W,\rho)$
\Comment{Alg.~\ref{alg:vertical-indices}}
\State $K,V \gets \textsc{Reorder}(K,V,\mathcal{I}_{\mathrm{VL}})$
\Comment{place vertical-line keys near the window}
\State Partition $[0,L_K)$ into $\mathcal{K}_{\mathrm{Q}}$, $\mathcal{K}_{\mathrm{VL}}$, and $\mathcal{K}_{\mathrm{Win}}$

\Statex \textcolor{vlblue}{\# Stage 2: quantize the quantizable prefix}
\For{each tile $T$ of size $B_N$ in $\mathcal{K}_{\mathrm{Q}}$}
    \State $s^K_T \gets \max(|K_T|)/c_B$
    \State $K^{\mathrm{Q}}_T \gets \mathrm{round}(K_T/s^K_T)$
    \Comment{$c_B=127$ for INT8; $c_B=448$ for FP8}
    \State Compute the query scale $s^Q$ for the corresponding query block
\EndFor

\Statex \textcolor{vlblue}{\# Stage 3: segmented online softmax}
\State $(m,\ell,acc) \gets (-\infty,0,\mathbf{0})$

\Statex \textcolor{vlblue}{\# Segment A: quantized prefix}
\For{each tile $T \subseteq \mathcal{K}_{\mathrm{Q}}$}
    \State $P_T \gets (Q^{\mathrm{Q}}_b {K^{\mathrm{Q}}_T}^{\top})(s^Q s^K_T)\alpha$
    \State $(m,\ell,acc) \gets \FlashOS(P_T,V_T;m,\ell,acc)$
\EndFor

\Statex \textcolor{vlblue}{\# Segment B: vertical-line tail}
\For{each tile $T \subseteq \mathcal{K}_{\mathrm{VL}}$}
    \State $P_T^{\mathrm{VL}} \gets Q_bK_T^{\top}\alpha$
    \State $(m,\ell,acc) \gets \FlashOS(P_T^{\mathrm{VL}},V_T;m,\ell,acc)$
\EndFor

\Statex \textcolor{vlblue}{\# Segment C: local full-precision window}
\For{each tile $T \subseteq \mathcal{K}_{\mathrm{Win}}$}
    \State Apply the per-row causal mask within $T$
    \State $P_T^{\mathrm{Win}} \gets Q_bK_T^{\top}\alpha$
    \State $(m,\ell,acc) \gets \FlashOS(P_T^{\mathrm{Win}},V_T;m,\ell,acc)$
\EndFor

\State \Return $O_b \gets acc/\ell$
\end{algorithmic}
\end{algorithm}


\subsection{Hybrid-Precision Decode Operator}
\label{app:decode_algorithm}

Algorithm~\ref{alg:hyquant_decode} describes the decode-stage attention kernel.
Decode is memory-bandwidth dominated, so HyQuant stores most historical KV states
in low-bit format and dequantizes them on the fly during attention computation.
The full-precision vertical-line tokens, staging buffer, and local window are
handled as separate segments and merged through the same online softmax state.
This avoids materializing a full-precision KV cache while preserving the most
error-sensitive tokens.

\begin{algorithm}[H]
\caption{\textbf{HyQuant Decode}: Hybrid-Precision KV-Cache Attention}
\label{alg:hyquant_decode}
\footnotesize
\begin{algorithmic}[1]
\Require Single-token query $q_t$ at step $t$
\Require KV buffers $(K^{\mathrm{Q}},V^{\mathrm{Q}})$, $(K^{\mathrm{VL}},V^{\mathrm{VL}})$,
$(K^{\mathrm{S}},V^{\mathrm{S}})$, $(K^{\mathrm{Win}},V^{\mathrm{Win}})$
\Require Bitwidths $(B_K,B_V)$; tile size $B_N$; split count $S$; GQA factor $g$; scale $\alpha=1/\sqrt{d}$
\Ensure Output $o_t$

\State $(m,\ell,acc) \gets (-\infty,0,\mathbf{0})$

\Statex \textcolor{vlblue}{\# Segment 1: quantized prefix}
\State Partition $K^{\mathrm{Q}}$ into $S$ chunks along the sequence axis
\For{$s=1,\ldots,S$ \textbf{ in parallel}}
    \State Dequantize $\hat{K}_s,\hat{V}_s$ from packed low-bit storage
    \State Pack $g$ GQA-grouped query heads into $\tilde{q}_t \in \mathbb{R}^{g \times d}$
    \State $P_s \gets \tilde{q}_t\hat{K}_s^{\top}\alpha$
    \State $(m_s,\ell_s,acc_s) \gets \FlashOS(P_s,\hat{V}_s;-\infty,0,\mathbf{0})$
\EndFor
\State $\mathcal{S}_{\mathrm{split}} \gets \{(m_s,\ell_s,acc_s)\}_{s=1}^{S}$
\State $(m,\ell,acc) \gets \ReduceSoftmax(\mathcal{S}_{\mathrm{split}})$

\Statex \textcolor{vlblue}{\# Segment 2: vertical-line tokens}
\If{$|K^{\mathrm{VL}}|>0$}
    \For{each tile $T$ in $K^{\mathrm{VL}}$}
        \State $P_T^{\mathrm{VL}} \gets q_t{K_T^{\mathrm{VL}}}^{\top}\alpha$
        \State $(m,\ell,acc) \gets \FlashOS(P_T^{\mathrm{VL}},V_T^{\mathrm{VL}};m,\ell,acc)$
    \EndFor
\EndIf

\Statex \textcolor{vlblue}{\# Segment 3: staging buffer}
\If{$|K^{\mathrm{S}}|>0$}
    \For{each tile $T$ in $K^{\mathrm{S}}$}
        \State $P_T^{\mathrm{S}} \gets q_t{K_T^{\mathrm{S}}}^{\top}\alpha$
        \State $(m,\ell,acc) \gets \FlashOS(P_T^{\mathrm{S}},V_T^{\mathrm{S}};m,\ell,acc)$
    \EndFor
\EndIf

\Statex \textcolor{vlblue}{\# Segment 4: local full-precision window}
\For{each tile $T$ in $K^{\mathrm{Win}}$}
    \State $P_T^{\mathrm{Win}} \gets q_t{K_T^{\mathrm{Win}}}^{\top}\alpha$
    \State $(m,\ell,acc) \gets \FlashOS(P_T^{\mathrm{Win}},V_T^{\mathrm{Win}};m,\ell,acc)$
\EndFor

\State \Return $o_t \gets acc/\ell$
\end{algorithmic}
\end{algorithm}


\subsection{Prefill-to-Decode KV-Cache Organization}
\label{app:freeze_algorithm}

Algorithm~\ref{alg:hyquant_freeze} describes how HyQuant reorganizes the KV
cache when switching from prefill to decode. The procedure fixes the
vertical-line token set selected from the prefill tail queries, keeps the recent
local window in full precision, and quantizes the remaining non-window prefix
into compact low-bit KV buffers. Newly generated tokens are first placed in a
small staging buffer before being merged into the hybrid layout. This design
avoids recomputing global token importance at every decode step, while keeping
the decode kernel structure simple: it scans the quantized prefix,
full-precision vertical-line tokens, staging states, and local window as four
separate segments.

\begin{algorithm}[H]
\caption{\textsc{Freeze}: Prefill$\to$Decode Hybrid KV-Cache Organization}
\label{alg:hyquant_freeze}
\footnotesize
\begin{algorithmic}[1]
\Require Prefill cache $K,V \in \mathbb{R}^{B \times H_{KV} \times N \times D}$
\Require Tail queries $Q_{\mathrm{tail}}$; window $W$; ratio $\rho$; bitwidths $(B_K,B_V)$; group size $G$
\Ensure Hybrid KV buffers for quantized prefix, vertical-line tokens, staging buffer, and local window

\Statex \textcolor{vlblue}{\# \VTI{} abbreviates \textsc{VerticalTokenIdentification}.}
\State $K^{\mathrm{Win}},V^{\mathrm{Win}} \gets K_{:,:,-W:,:},V_{:,:,-W:,:}$
\State $\mathcal{I}_{\mathrm{VL}} \gets \VTI(Q_{\mathrm{tail}},K,W,\rho)$
\Comment{computed once and held fixed}
\State $K^{\mathrm{VL}},V^{\mathrm{VL}} \gets K[:,:,\mathcal{I}_{\mathrm{VL}},:],V[:,:,\mathcal{I}_{\mathrm{VL}},:]$
\State $\mathcal{I}_{\mathrm{Q}} \gets [0,N-W)\setminus \mathcal{I}_{\mathrm{VL}}$
\State $K^{\mathrm{Q}} \gets \mathrm{Quant}_{B_K}(K[:,:,\mathcal{I}_{\mathrm{Q}},:];G)$
\State $V^{\mathrm{Q}} \gets \mathrm{Quant}_{B_V}(V[:,:,\mathcal{I}_{\mathrm{Q}},:];G)$
\State Initialize empty staging buffer $K^{\mathrm{S}},V^{\mathrm{S}}$ with capacity $W_s$
\State Free $K,V$
\State \Return all four buffers
\end{algorithmic}
\end{algorithm}

\FloatBarrier


\section{Additional Experimental Results}
\label{app:additional_results}

This appendix reports additional ablations and auxiliary evaluations supporting
Section~\ref{sec:experiments}. Unless otherwise stated, results use Qwen3-8B
with K4V4 quantization and vertical-line-aware full-precision retention.

\subsection{Math Reasoning Results}
\label{app:math_reasoning}

We additionally report math reasoning accuracy on GSM8K and MATH500. These
results complement the LongBench evaluation in the main text.

\begin{table}[H]
  \centering
  \small
  \setlength{\tabcolsep}{4pt}
  \renewcommand{\arraystretch}{1.12}
  \caption{\textbf{Math reasoning accuracy on Qwen3-8B.}}
  \label{tab:math_main_qwen}
  \begin{tabular*}{0.92\columnwidth}{@{\extracolsep{\fill}}lcc}
    \toprule
    \textbf{Method} & \textbf{GSM8K} & \textbf{MATH500} \\
    \midrule
    FA2              & 95.88 & \textbf{80.14} \\
    KIVI             & 92.48 & 72.89 \\
    KVTuner          & 93.69 & 75.83 \\
    SageAttention    & 94.99 & 77.60 \\
    \textbf{HyQuant} & \textbf{96.52} & 78.73 \\
    \bottomrule
  \end{tabular*}
\end{table}


\subsection{Sensitivity to the Local Full-Precision Window}
\label{app:window_ablation}

We vary the number of recent tokens retained in full precision. A larger local
window slightly improves accuracy and reduces attention-output MSE.

\begin{table}[H]
  \centering
  \small
  \setlength{\tabcolsep}{4pt}
  \renewcommand{\arraystretch}{1.12}
  \caption{\textbf{Sensitivity to the local full-precision window size.}
  MSE is reported in units of $10^{-2}$; lower is better.}
  \label{tab:ablation_win}
  \begin{tabular*}{0.92\columnwidth}{@{\extracolsep{\fill}}ccc}
    \toprule
    \textbf{Window} & \textbf{Acc.} & \textbf{MSE} \\
    \midrule
    64  & 92.54 & 9.58 \\
    128 & 95.52 & 8.48 \\
    256 & \textbf{95.74} & \textbf{8.21} \\
    \bottomrule
  \end{tabular*}
\end{table}

\begin{table*}[!t]
  \centering
  \small
  \setlength{\tabcolsep}{4.6pt}
  \renewcommand{\arraystretch}{1.12}
  \caption{\textbf{Sensitivity to the retained vertical-line token ratio.}
  LongBench subset results on Qwen3-8B under K4V4 quantization.}
  \label{tab:ablation_k}
  \begin{tabular*}{0.98\textwidth}{@{\extracolsep{\fill}}lcccccccc}
    \toprule
    \textbf{Top-$k$} &
    \textbf{NarrQA} &
    \textbf{Qasper} &
    \textbf{MF-en} &
    \textbf{HotpotQA} &
    \textbf{Musique} &
    \textbf{SAMSum} &
    \textbf{Pretrieve-en} &
    \textbf{Avg.} \\
    \midrule
    2\%  & 24.24 & 38.69 & 45.04 & 47.50 & 35.25 & 35.49 & 96.19  & 46.06 \\
    4\%  & 23.20 & 40.72 & 44.47 & 53.40 & 43.23 & 35.59 & 98.82  & 48.49 \\
    6\%  & 23.85 & 37.67 & \textbf{53.38} & \textbf{62.65} & 41.98 & 36.52 & 99.26  & 50.76 \\
    8\%  & 23.65 & 41.84 & 48.87 & 57.96 & 36.53 & 37.26 & 95.23  & 48.76 \\
    10\% & \textbf{26.79} & \textbf{45.73} & 49.61 & 56.23 & \textbf{45.40} & \textbf{39.61} & \textbf{100.00} & \textbf{51.91} \\
    \bottomrule
  \end{tabular*}
\end{table*}

\begin{table*}[t]
\centering
\small
\caption{MInference-inspired vertical-only ablation on Qwen3-8B.
Metric is F1 for narrativeqa, hotpotqa, triviaqa and ROUGE-L for multi\_news. Each dataset uses the first 100 test cases.}
\label{tab:minference-concept-ablation}
\begin{tabular}{lcccccc}
\toprule
Task & Metric & FA2 & MInf.-Vertical & HyQuant
     & $\Delta$(Mv.\,$-$\,FA2) & $\Delta$(HQ.\,$-$\,Mv.) \\
\midrule
narrativeqa& F1      & 31.49 & 14.69 & 31.49 & $-16.80$ & $+16.80$ \\
\addlinespace[2pt]
hotpotqa & F1      & 62.83 & 38.55 & 62.83 & $-24.28$ & $+24.28$ \\
\addlinespace[2pt]
multi\_news        & ROUGE-L & 14.45 & 13.87 & 14.52 & $-0.59$  & $+0.65$  \\
\addlinespace[2pt]
triviaqa & F1      & 89.86 & 79.20 & 91.25 & $-10.66$ & $+12.05$ \\
\midrule
average &         & 49.66 & 36.58 & 49.97 & $-13.08$ & $+13.39$ \\
\bottomrule
\end{tabular}
\end{table*}

\subsection{Sensitivity to the Vertical-Line Token Ratio}
\label{app:ratio_ablation}

Table~\ref{tab:ablation_k} reports the effect of changing the retained
vertical-line token ratio. Increasing the ratio generally improves the average
score, but it also increases the number of full-precision KV states. We use
top-$5\%$ in the main experiments as a practical accuracy--efficiency trade-off.


\subsection{Short-Context Evaluation}
\label{app:short_context}

Although HyQuant is designed for long-context inference, we also evaluate it on
short-context benchmarks. HyQuant remains competitive with the full-precision
baseline while substantially outperforming strict 4-bit KIVI.

\begin{table}[H]
  \centering
  \small
  \setlength{\tabcolsep}{3pt}
  \renewcommand{\arraystretch}{1.12}
  \caption{\textbf{Short-context evaluation on Qwen3-8B.}}
  \label{tab:short_context_qwen}
  \begin{tabular*}{0.92\columnwidth}{@{\extracolsep{\fill}}lcccc}
    \toprule
    \textbf{Method} & \textbf{C-Eval} & \textbf{MMLU} & \textbf{GSM8K} & \textbf{Avg.} \\
    \midrule
    FA2              & \textbf{85.32} & \textbf{86.45} & 94.89 & \textbf{88.89} \\
    KIVI-K8V8        & 82.83 & 84.77 & 94.42 & 87.34 \\
    KIVI-K4V4        & 75.26 & 78.34 & 91.88 & 81.83 \\
    \textbf{HyQuant} & 83.23 & 84.79 & \textbf{95.43} & 87.82 \\
    \bottomrule
  \end{tabular*}
\end{table}


\subsection{Component-Level MSE Ablation}
\label{app:mse_ablation}

We further isolate the effect of local-window retention and vertical-line
retention. The combination of both components gives the lowest layer-wise MSE.

\begin{figure}[H]
  \centering
  \includegraphics[width=0.95\columnwidth]{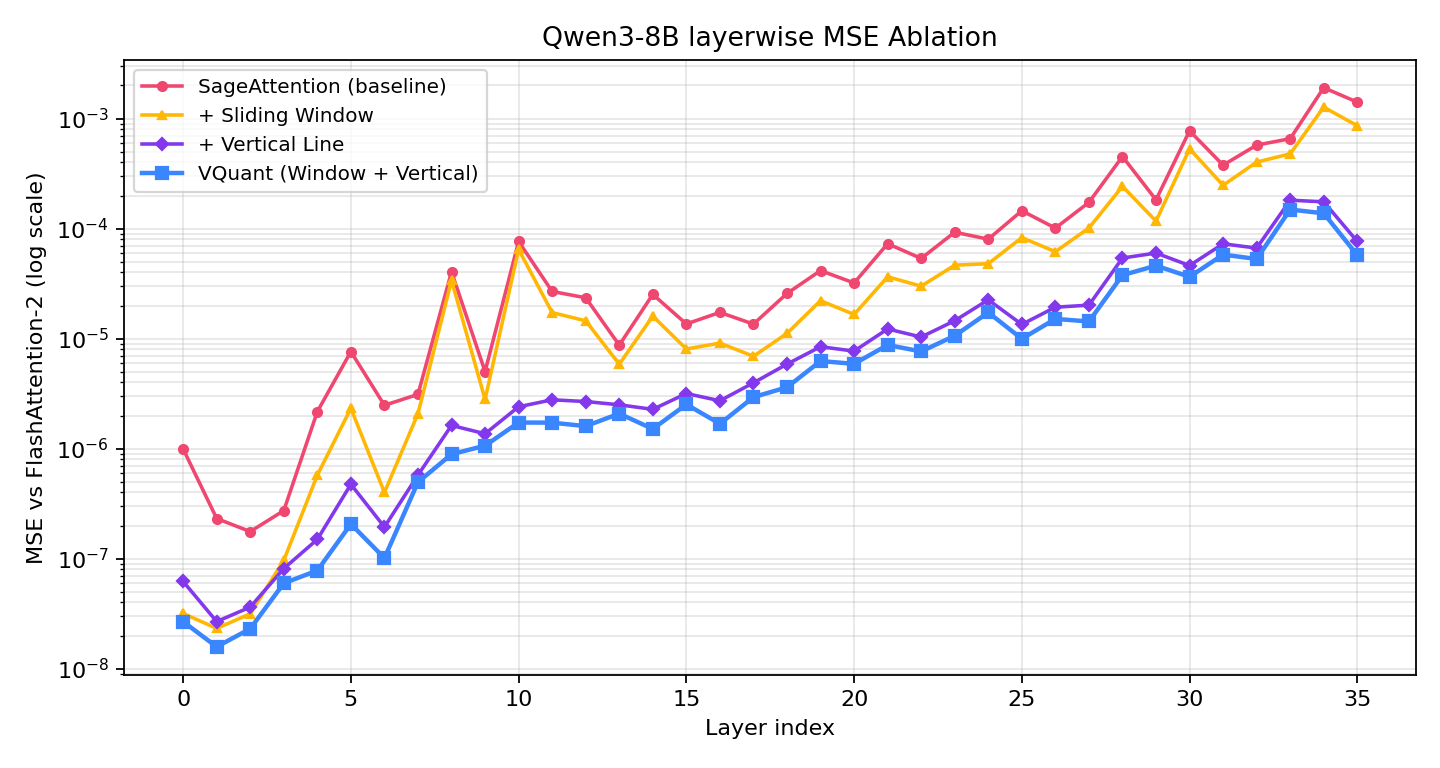}
  \caption{\textbf{Component-level MSE ablation of HyQuant.}
  We compare uniform quantization, local-window retention, vertical-line
  retention, and their combination on Qwen3-8B. Lower MSE is better.}
  \label{fig:ablation_mse}
\end{figure}

\subsection{Conceptual ablation: marginal value of the quantized long tail} 
\label{sec:conceptablation}
MInference~\cite{jiang2024minference10acceleratingprefilling} accelerates prefill by computing attention only over dynamically selected sparse positions, including its Vertical-Slash pattern. A vertical-only sparse variant may omit useful long-tail positions in long-context settings, whereas HyQuant retains these positions in low precision.
To measure the marginal effect of retaining the long tail, we conduct an ablation study shown in Table~\ref{tab:minference-concept-ablation}. The results demonstrate the benefit of keeping all tokens in HyQuant: non-vertical positions remain available in low precision, distinguishing our mixed-precision framework from sparse attention.
\subsection{Memory overhead breakdown}
\label{app:memory_overhead}
Table \ref{tab:memory-overhead} provides memory overhead analysis for HyQuant.
\begin{table}[t]
\centering
\small
\begin{tabular}{lrr}
\toprule
\textbf{Component} & \textbf{8K prefix} & \textbf{32K prefix} \\
\midrule
VL tokens (5\% of prefix)    & $+15.0\%$ & $+15.0\%$ \\
Local window ($W{=}128$)     & $+4.7\%$  & $+1.2\%$  \\
Staging buffer ($W_s{=}128$) & $+4.7\%$  & $+1.2\%$  \\
Query buffer (VTI)           & $<0.1\%$  & $<0.1\%$  \\
\midrule
\textbf{Total} & $\sim$24.4\% & $\sim$17.4\% \\
\bottomrule
\end{tabular}
\caption{Memory overhead breakdown vs.\ strict K4V4 (Qwen3-8B, $W{=}128$, $\rho{=}5\%$, per layer).}
\label{tab:memory-overhead}
\end{table}

\subsection{Experimental Configuration Details}
\label{app:exp_config}

Table~\ref{tab:exp_config} summarizes the common experimental configuration used across all methods compared in this paper, including hardware, kernel source, batch size, prefill and generation length, KV bit-width, and memory budget.

\begin{table}[h]
\centering
\small
\caption{Experimental configuration used for all compared methods.}
\label{tab:exp_config}
\begin{tabular}{ll}
\toprule
\textbf{Setting} & \textbf{Configuration (all methods)} \\
\midrule
Hardware & NVIDIA H100 (80GB) \\
Batch size & 1 \\
Prefill length & Up to 128K tokens \\
Generation length & Up to 32K tokens \\
KV bit-width & 4-bit for all quantization methods \\
Memory budget & Unconstrained;  \\
Window size & 128 \\
\bottomrule
\end{tabular}
\end{table}

\FloatBarrier

\end{document}